\documentclass[lettersize,journal]{IEEEtran}
\usepackage{cite}
\usepackage{amsmath,amsfonts}
\usepackage{algorithmic}
\pdfoutput=1
\usepackage{array}
\usepackage[caption=false,font=normalsize,labelfont=sf,textfont=sf]{subfig}
\usepackage{textcomp}
\usepackage{stfloats}
\usepackage{url}
\usepackage{verbatim}
\usepackage{graphicx}
\usepackage{color}
\usepackage{hyperref}
\hypersetup{
    colorlinks=true,             
    linkcolor=blue,              
    citecolor=blue,              
    filecolor=mycustompurple,    
    urlcolor=magenta             
}
\usepackage{multirow}
\usepackage{booktabs} 
\usepackage{bm}

\def\BibTeX{{\rm B\kern-.05em{\sc i\kern-.025em b}\kern-.08em
    T\kern-.1667em\lower.7ex\hbox{E}\kern-.125emX}}
\usepackage{balance}
\begin{document}

\title{ A Complex-valued SAR Foundation Model Based on Physically Inspired Representation Learning}


\author{
 Mengyu~Wang,  
 Hanbo~Bi, 
 Yingchao~Feng,
 Linlin~Xin,
 Shuo~Gong,
 Tianqi~Wang,\\
 Zhiyuan~Yan,
 Peijin~Wang,
 Wenhui~Diao,
 and~Xian~Sun,~\IEEEmembership{Senior Member,~IEEE}
        
\thanks{This work was supported by the National Natural Science Foundation of China (NSFC) under Grant 62301538. \textit{(M. Wang, H. Bi and Y. Feng contribute equally to this work.) (Corresponding author: W. Diao.)}}

\thanks{M. Wang, H. Bi, L. Xin, S. Gong, T. Wang, P. Wang, W. Diao and X. Sun are with the Aerospace Information Research Institute, Chinese Academy of Sciences, Beijing 100190, China, also with the School of Electronic, Electrical and Communication Engineering, University of Chinese Academy of Sciences, Beijing 100190, China, also with the University of Chinese Academy of Sciences, Beijing 100190, China, and also with the Key Laboratory of Target Cognition and Application Technology (TCAT), Aerospace Information Research Institute, Chinese Academy of Sciences, Beijing 100190, China (e-mail: wangmengyu22@mails.ucas.edu.cn; bihanbo21@mails.ucas.edu.cn).}
 
\thanks{Y. Feng and Z. Yan are with the Aerospace Information Research Institute, Chinese Academy of Sciences, Beijing 100190, China, and also with the Key Laboratory of Target Cognition and Application Technology (TCAT), Aerospace Information Research Institute, Chinese Academy of Sciences, Beijing 100094, China (e-mail: fengyc@aircas.ac.cn).}

}

\markboth{Journal of \LaTeX\ Class Files,~Vol.~18, No.~9, September~2020}%
{How to Use the IEEEtran \LaTeX \ Templates}

\maketitle

\begin{abstract}
Vision foundation models in remote sensing have been extensively studied due to their superior generalization on various downstream tasks. Synthetic Aperture Radar (SAR) offers all-day, all-weather imaging capabilities, providing significant advantages for Earth observation. However, establishing a foundation model for SAR image interpretation inevitably encounters the challenges of insufficient information utilization and poor interpretability. 
In this paper, we propose a remote sensing foundation model based on complex-valued SAR data, which simulates the polarimetric decomposition process for pre-training, i.e., characterizing pixel scattering intensity as a weighted combination of scattering bases and scattering coefficients, thereby endowing the foundation model with physical interpretability.
Specifically, we construct a series of scattering queries, each representing an independent and meaningful scattering basis, which interact with SAR features in the scattering query decoder and output the corresponding scattering coefficient.
To guide the pre-training process, polarimetric decomposition loss and power self-supervision loss are constructed. The former aligns the predicted coefficients with Yamaguchi coefficients, while the latter reconstructs power from the predicted coefficients and compares it to the input image's power.
The performance of our foundation model is validated on six typical downstream tasks, achieving state-of-the-art results. Notably, the foundation model can extract stable feature representations and exhibits strong generalization, even in data-scarce conditions.
\end{abstract}



\section{Introduction}

\begin{figure}[!t]
\centering
\setlength{\abovecaptionskip}{1pt}
\setlength{\belowcaptionskip}{1pt}
\includegraphics[width=1.0\linewidth]{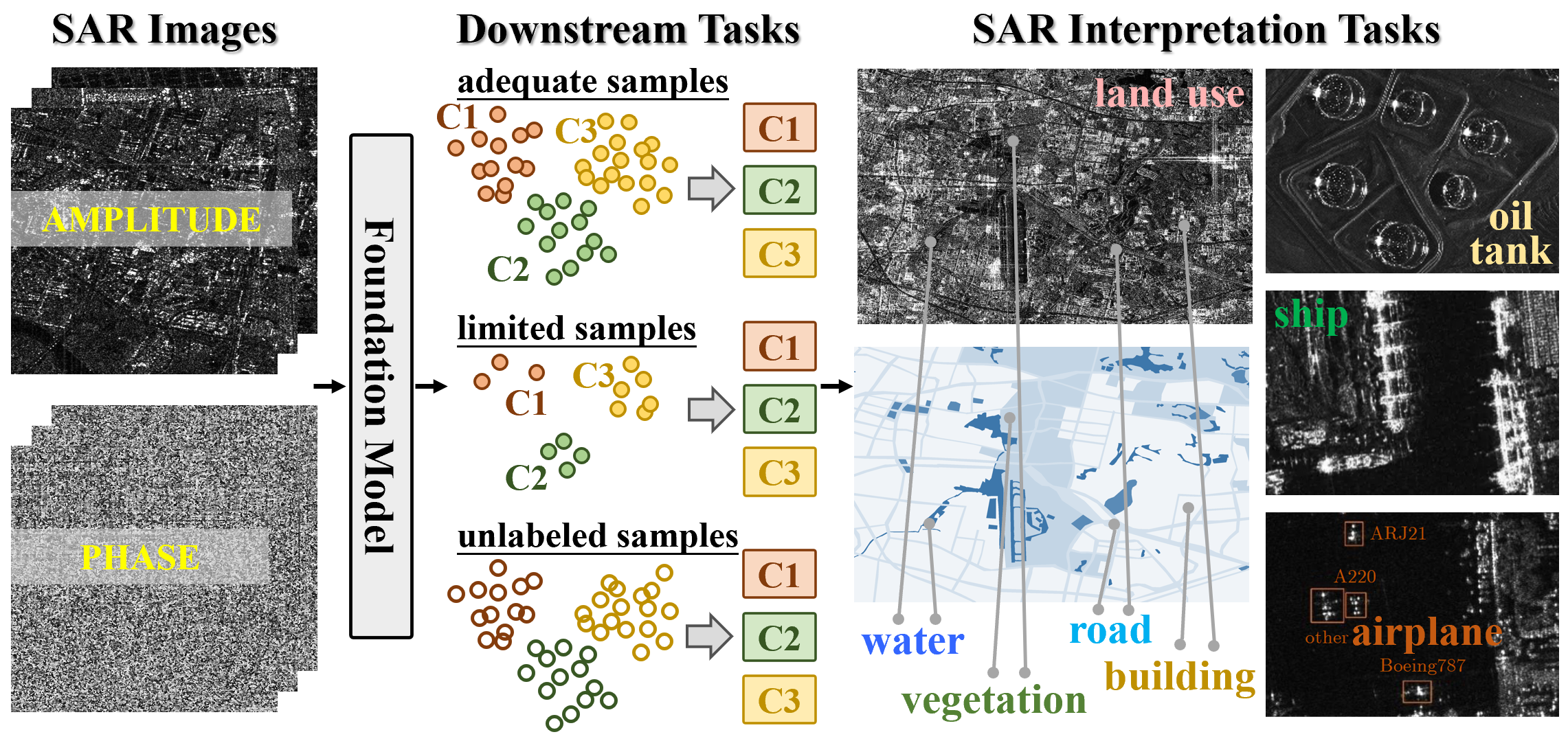}
\caption{\textbf{The paradigm of SAR foundation model + downstream tasks.} The foundation model explores unlabeled SAR data to extract the semantics of interested areas and classes, and the pre-trained foundation model can be broadly applied across various downstream tasks and applications.}
\label{SARapplication}
\end{figure}

The number of remote sensing (RS) satellites launched has increased significantly, leading to the generation of vast amounts of RS imagery. To maximize the potential of this data, the development of foundation models for RS has emerged as a prominent research focus~\cite{RSpretrained, RVSA, RSpretrained_b}. These models aim to learn universal representations from unlabeled data by employing self-supervised pre-training paradigms, such as masked image modeling~\cite{Simmim} and contrastive learning~\cite{wang2021dense}. As illustrated in Fig.~\ref{SARapplication}, pre-trained foundation models exhibit strong generalization capabilities and can be effectively applied to various downstream tasks, including scene classification, object detection, and semantic segmentation.

In recent years, several RS foundation models have been introduced. For instance, Sun et al. developed RingMo\cite{RSpretrained2}, the first RS foundation model (RSFM) pre-trained on two million optical images. Subsequent efforts, such as Scale-MAE\cite{Scale-mae} and RVSA\cite{RVSA}, scaled up both pre-training datasets and model parameters, achieving further improvements in generalization performance. However, despite these advancements, most RSFMs focus predominantly on optical data. There remains limited progress in extending foundation models to other modalities of RS data, such as spectral~\cite{SpectralGPT}, synthetic aperture radar (SAR)~\cite{SARATR-X, SPT}, and multi-modal imagery~\cite{CROMA}. Among these, SAR presents unique opportunities and challenges that demand tailored solutions.

SAR is an active microwave imaging sensor with significant advantages over other RS modalities~\cite{zhu2021deep}. Unlike optical sensors, SAR operates independently of sunlight, weather, or time of day, enabling consistent acquisition of high-resolution images under all conditions. Its use of microwave frequencies allows SAR to penetrate clouds, fog, and vegetation, making it highly effective for capturing hydrological features, structural details, and other critical information~\cite{MCANet, DENet}. Despite its strengths, traditional interpretation methods for SAR data are insufficient to fully exploit its potential. There is an urgent need for foundation models specifically designed to address the distinct characteristics and challenges of SAR data.

\begin{figure}[!t]
\centering
\setlength{\abovecaptionskip}{1pt}
\setlength{\belowcaptionskip}{1pt}
\includegraphics[width=1.0\linewidth]{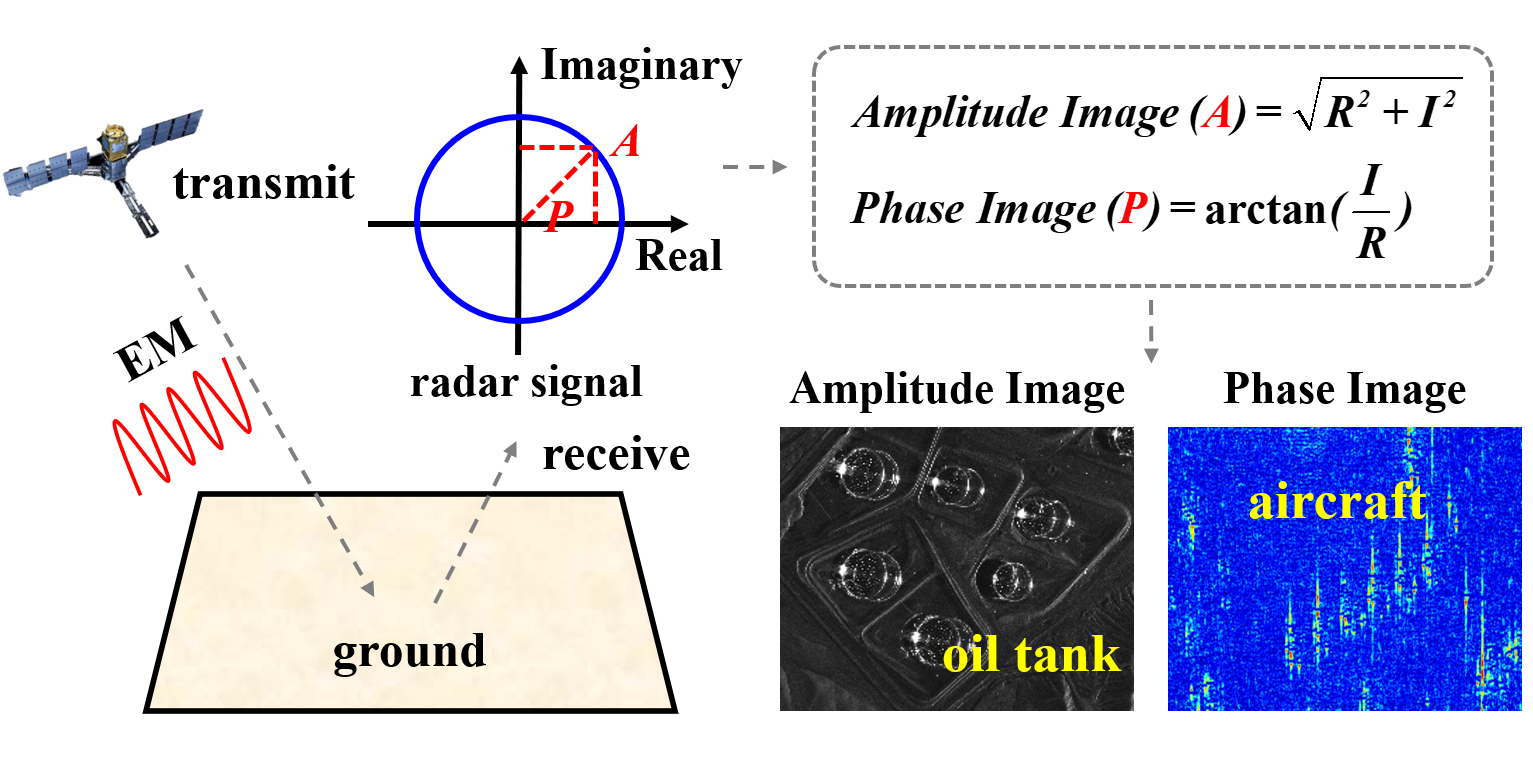}
\caption{\textbf{Amplitude and phase information in complex-valued SAR imagery.} The radar sensor transmits electromagnetic (EM) waves and receives complex-valued radar signals. The amplitude image highlights strong scattering points and clear contours, while the phase image contains more detailed information about the structure of objects, such as aircraft.}
\label{AmplitudePhase}
\end{figure}

First, fully utilizing the information inherent in SAR images is essential for improving model performance. As illustrated in Fig.\ref{AmplitudePhase}, most existing SAR research focuses on amplitude-based single-channel images, which capture only magnitude information. These images reflect radar echo intensity but disregard phase information—an integral component of SAR’s original complex-valued format. Phase data encodes critical physical information, such as the distance between radar waves and targets\cite{9780199}. Incorporating phase information into foundation models would enable the learning of more stable and comprehensive radar representations, significantly enhancing SAR’s capacity for precise and efficient Earth observation.

Second, model adaptability and interpretability pose additional challenges. Unlike optical images, which emphasize texture and color, SAR imagery primarily captures scattering characteristics. Consequently, existing self-supervised pretraining methods designed for pixel RGB reconstruction are poorly suited for SAR, as they fail to effectively represent its scattering properties. Furthermore, current encoder-decoder architectures often produce latent feature embeddings that lack physical interpretability~\cite{li2022interpretable}. These limitations reduce the ability of foundation models to reliably represent and analyze SAR imagery.

To overcome these challenges, endowing deep networks with physical interpretability is a promising solution. Within the context of SAR imaging, the scattering intensity of a single pixel can be mathematically expressed as a weighted combination of various scattering mechanisms (e.g., surface scattering, double-bounce scattering) via polarimetric decomposition~\cite{Yamaguchi2005four}. That is, this process characterizes the scattering characteristics of SAR by defining specific scattering bases and their corresponding coefficients. Inspired by this, we model and leverage these scattering mechanisms to construct interpretable radar feature representations. By effectively capturing the unique scattering characteristics of SAR imagery, such a paradigm offers a robust framework for SAR data interpretation. 

\begin{figure}[t]
\centering
\setlength{\abovecaptionskip}{1pt}
\setlength{\belowcaptionskip}{1pt}
\includegraphics[width=1.0\linewidth]{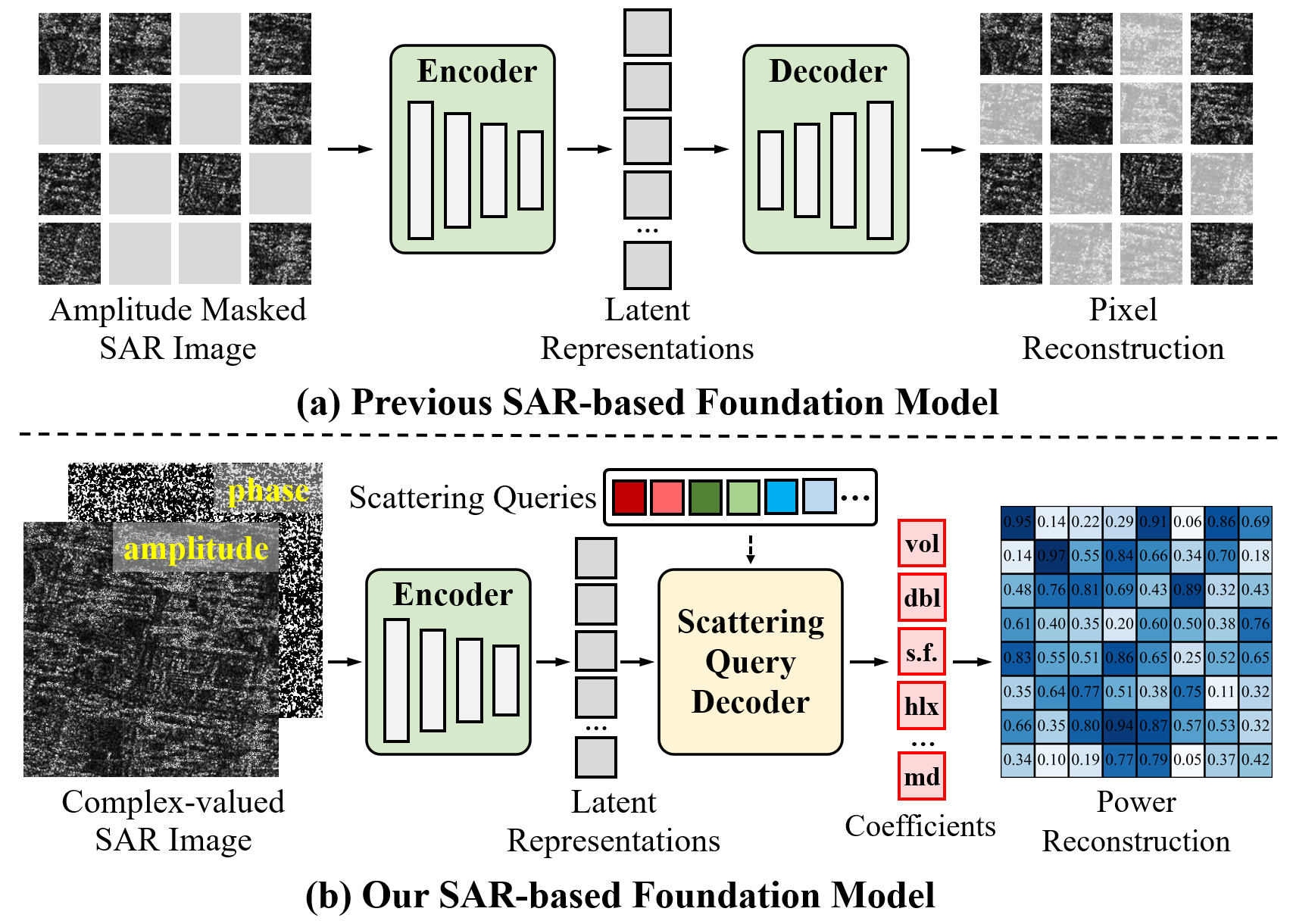}
\caption{\textbf{Comparison of previous SAR-based foundation model and ours.} (a) Pevious SAR-based foundation models use amplitude images as input and reconstruct pixel values through masked image modeling\cite{CROMA}. (b) Our foundation model takes amplitude and phase images as input and reconstructs power values by simulating the physical process of SAR polarimetric decomposition.}
\label{intro}
\end{figure}


Thus, we develop a self-supervised foundation model by simulating the physical process of SAR polarimetric decomposition, which is different from the previous RSFM pre-training methods, as shown in Fig. \ref{intro}. 
Previous SAR-based foundation models take the amplitude SAR image as input, extract latent feature representations, and reconstruct the corresponding pixel values\cite{CROMA}. 
Instead, our model utilizes a polarization pixel encoder to extract SAR features from complex-valued SAR data. 
To simulate the scattering bases in the decomposition process, we construct scattering queries (i.e. feature embeddings), each of which is independent and physically meaningful. Specifically, the scattering bases in matrix form are vectorized to initialize the scattering queries. Then the scattering queries interact with SAR features via a dedicated scattering query decoder. Such a process simulates the polarimetric decomposition and generates the scattering coefficients corresponding to the scattering bases. 
To guide the pre-training of the foundation model on a large amount of unlabeled data, the polarimetric decomposition loss and power self-supervised loss are formulated. Specifically, polarimetric decomposition loss enables the network to enforce supervised constraints on the four-component coefficients (volume scattering, double-bounce scattering, surface scattering, helix scattering) obtained through the Yamaguchi algorithm~\cite{Yamaguchi2005four} and network predictions. Meanwhile, leveraging the principle that the total power of various scattering components equals the image power~\cite{cloude1996}, a power adapter module is designed to reconstruct the image power from the predicted scattering coefficients and perform power self-supervised loss.


To validate the superiority and generalization of our foundation model, we conduct a comprehensive evaluation on six tasks across complex-valued SAR datasets (semantic segmentation, few-shot segmentation, and unsupervised classification) and general SAR datasets (ship detection, aircraft detection, and semantic segmentation). Our model consistently achieves state-of-the-art performance across all tasks, demonstrating competitive results even under few-shot and unsupervised conditions.

In summary, the contributions of this paper can be summarised as follows:
\begin{enumerate}
    \item For the first time, a remote sensing foundation model based on complex-valued SAR data is proposed. This foundation model is pre-trained by simulating the physical process of polarimetric decomposition, which represents pixel scattering intensity as a weighted combination of scattering bases and coefficients, thereby endowing the model with physical interpretability.
    \item We construct scattering queries, a set of physically meaningful feature embeddings to represent the scattering bases. These scattering queries interact with SAR features in the scattering query decoder to simulate the polarimetric decomposition process and output scattering coefficients.
    \item To guide the pre-training process, polarimetric decomposition loss and power self-supervision loss are formulated. The former uses the predicted coefficients and Yamaguchi coefficients to perform loss, while the latter reconstructs power from predicted coefficients and compares it with the input image's power.
    \item Extensive experiments on six tasks across complex-valued SAR datasets and general SAR datasets show that our foundation model achieves state-of-the-art performance and has strong generalization, even under data-scarce conditions.
    
\end{enumerate}

\section{Related Work}
\subsection{Remote Sensing Foundation Models}
Recently, foundation models have become one of the hot topics in artificial intelligence. These models are pre-trained on extensive datasets using self-supervised learning and exhibit superior performance and powerful generalization across various downstream tasks\cite{Llama,MAE,SAM}. Researchers in the RS community have employed this paradigm to develop foundation models for a variety of RS data.

For optical RS images,  Sun et al. introduce the first RS foundation model, RingMo\cite{RSpretrained2}, which optimizes the masking strategy to focus on dense and small objects. RVSA\cite{RVSA} proposes a rotated window attention mechanism, which better copes with large-scale and arbitrarily oriented objects in RS images. Scale-MAE\cite{Scale-mae} reconstructs high-frequency and low-frequency image representations at different scales and improve the multi-scale perception capabilities of the foundation model. 
Furthermore, expanding the RGB-band to multi-band RS images, Cong et al.\cite{SatMAE} encode multi-spectral data as groups of bands with spectral position embedding and propose multi-spectral RS foundation model. SpectralGPT\cite{SpectralGPT} proposes 3D token generation for spatial-spectral coupling to better capture the continuity of multi-spectral data.

Given the unique advantages of SAR imaging under all-day and all-weather conditions, researchers have devoted their attention to explore SAR foundation models\cite{SARATR-X,FGMAE,DINO-MM,CROMA,guo2024skysense}. CROMA\cite{CROMA} performs mask reconstruction and cross-modal contrastive learning on SAR and multi-spectral optical samples, exploring the potential of SAR data and demonstrating advantages in inferring large-scale images. SARATR-X\cite{SARATR-X} systematically investigates the construction of a SAR ATR foundation model and achieves strong generalization under multi-target, multi-scene, and multi-sensor conditions. FG-MAE\cite{FGMAE} improves the reconstruction strategy by replacing the raw image with a manually designed histogram of oriented graidents feature, thereby enhancing the spatial information and reducing the impact of speckle noise. However, domain differences between optical and SAR images necessitate further research to enhance SAR image understanding.

\subsection{Polarimetric Decomposition}
Polarimetric decomposition analysis in SAR imagery represents a key area within RS. It derives the polarization characteristics of targets from PolSAR data, enabling enhanced comprehension and precise recognition of terrestrial features.
Current methodologies are mainly classified into two types: coherent decomposition and incoherent decomposition\cite{PolarConvNetwork}.

Coherent decomposition decomposes the single target scattering matrix into the form of basic target scattering components to infer the possible physical structure of the target. The commonly used coherent decomposition models include Pauli decomposition\cite{PauliDecomposition}, Krogager decomposition\cite{KrogagerDecomposition}, Cameron decomposition\cite{CameronDeconposition}, Touzi decomposition\cite{TouziDecomposition}, and some advances\cite{ISTDDecomposition, chen2018advanced}. The advantage of coherent decomposition lies in its ability to represent any scattering matrix $S$ as the summation of scattering matrices associated with basic targets. These basic targets can be linked to specific deterministic scattering mechanisms. 

However, in practical RS applications, the information captured by a single pixel typically involves contributions from multiple scatterers. This situation introduces the need for incoherent decomposition techniques. Incoherent decomposition breaks down the Kennaugh matrix, coherence matrix, or covariance matrix of a distributed target into a combination of second-order descriptors and provides a physical explanation. Huynen decomposition\cite{Huynen1970} lays the foundation for the field of polarimetric decomposition. Some classic methods include Freeman-Durden decomposition\cite{Freeman-DurdenDecomposition}, Yamaguchi four-component decomposition\cite{Yamaguchi2005four}, and Cloud decomposition\cite{cloude1996}.
Among these techniques, Yamaguchi four-component decomposition has garnered significant public recognition. Furthermore, a number of model-based decomposition techniques such as seven-component decomposition\cite{C7D_2019, C7D_2023} and eight-component decomposition\cite{C8D_2022} have been developed and refined based on the principles established by Yamaguchi four-component decomposition.

\section{Preliminary: Physical Decomposition Process}
This section introduces the physical process of polarimetric decomposition of SAR. The SAR sensor transmits and receives horizontal (H) and vertical (V) polarized electromagnetic waves to obtain SAR images in four modes: HH, HV, VH, and VV, namely fully polarized SAR (PolSAR) images. Each pixel unit is in a complex-valued data format, and the amplitude and phase information that can describe the target scattering characteristics are obtained\cite{C7D_2023, C8D_2022}.

For coherent targets, there exists a linear transformation relationship between radar irradiation waves and target scattering waves, which can be represented by a complex-valued two-dimensional polarization scattering matrix $S$.
The received scattering wave is expressed as:
\begin{equation}
E^s=S E^i=\frac{e^{-i k r}}{r}\left(\begin{array}{ll}
S_{\mathrm{HH}} & S_{\mathrm{HV}} \\
S_{\mathrm{VH}} & S_{\mathrm{VV}}
\end{array}\right)\binom{E_{\mathrm{H}}^i}{E_{\mathrm{V}}^i}
\end{equation}
where $i$ and $s$ represent the incident wave and the received scattering wave respectively, $r$ is the distance between the target and the antenna, and $k$ is the number of electromagnetic waves. 

\begin{figure}[t]
\centering
\setlength{\abovecaptionskip}{1pt}
\setlength{\belowcaptionskip}{1pt}
\includegraphics[width=1.0\linewidth]{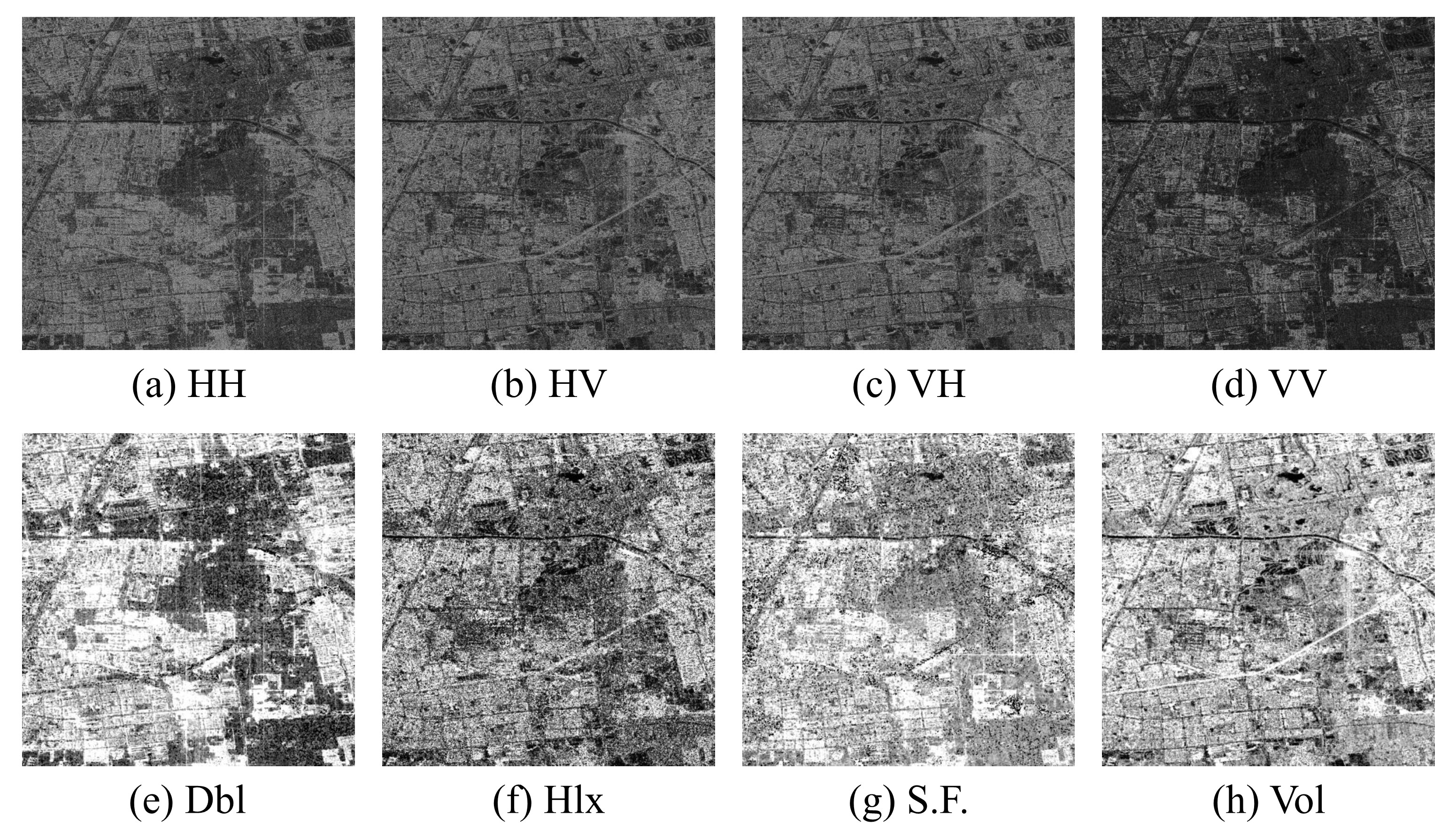}
\caption{\textbf{Schematic diagram of the Yamaguchi decomposition components.} 
The top line exhibits fully PolSAR images, arranged from left to right with polarization modes HH, HV, VH, and VV respectively. The subsequent line depicts the results of four components from $T$, illustrating Dbl scattering, Hlx scattering, S.F. scattering, and Vol scattering components respectively.}
\label{physical_process_of_sar_imaging}
\end{figure}

In the practical application of PolSAR, the assumption of pure coherent scatterers is not valid. The information recorded in a single pixel contains multiple scatterers, and the $S$ value of a single pixel is the coherent superposition of the $S$ matrix of all scatterers in the resolution unit. In the widely used model-based incoherent decomposition method\cite{Yamaguchi2005four}, the distributed target is described using polarization coherence matrix $T$ derived from $S$. The polarization scattering vector is expressed as:




\begin{equation}
k_p=\frac{1}{\sqrt{2}}\left[S_{\mathrm{HH}}+S_{\mathrm{HV}}, S_{\mathrm{VV}}-S_{\mathrm{HH}}, 2 S_{\mathrm{HV}}\right]^T
\end{equation}
Then, the coherence matrix $T$ can be created from $k_p$ as:
\begin{equation}
T=<k_p k_p^H>=
\left[\begin{array}{lll}
T_{11}     &T_{12}      &T_{13}  \\
T_{12}^{*} &T_{22}      &T_{23}  \\
T_{13}^{*} &T_{23}^{*}  &T_{33}
\end{array}\right]
\end{equation}
where $H$ denotes the complex conjugation and transposition, $<\cdot>$ represents the spatial average. Polarimetric decomposition decomposes the scattering of ground objects into the weighted sum of multiple scattering components. Taking Yamaguchi decomposition as an example, the formula used is as follows,
\begin{equation}
[T] = P_\mathrm{d}[T]_\mathrm{d} + P_\mathrm{h}[T]_\mathrm{h} + P_\mathrm{s}[T]_\mathrm{s} + P_\mathrm{v}[T]_\mathrm{v}
\end{equation}
where the subscripts d, h, s and v represent double-bounce (Dbl) scattering, helix (Hlx) scattering, surface (S.F.) scattering, and volume (Vol) scattering components respectively. The schematic diagram of the physical process of polarimetric decomposition is shown in Fig. \ref{physical_process_of_sar_imaging}.

\section{Methodology}
This section offers an in-depth introduction to our physically inspired foundation model, which focuses on complex-valued SAR data. Firstly, the design concept and overview are presented in Sec.\ref{overiew}, followed by an outline of the network framework in Sec.\ref{nframework}. The Sec.\ref{sq} and Sec.\ref{loss} introduce scattering query construction and self-supervised loss respectively.

\subsection{Overview} \label{overiew}
Direct visual interpretation of SAR images is highly challenging. We consider constructing a foundation model from the perspective of physical interpretability to enhance its generalization and interpretation performance in various downstream tasks. The physical process of SAR polarimetric decomposition shows that the proportions of different scattering components (such as Dbl, Hlx, S.F., and Vol) vary for each pixel, that is, the process is characterized as a weighted combination of scattering bases and scattering coefficients. Inspired by this, we propose, for the first time, a self-supervised pre-trained foundation model based on the physical process of polarimetric decomposition from complex-valued SAR data, effectively leveraging the amplitude, phase, and polarization information inherent in these images.

To simulate the physical process of polarimetric decomposition using deep networks, three main issues must be addressed. The first step is to connect the physical process with the deep network. Polarimetric decomposition can be viewed as a combination of a series of scattering bases and corresponding coefficients. In mathematical models, the Taylor formula embodies a similar concept, expressing function values as combinations of polynomial bases and polynomial coefficients. Considering that the essence of deep networks lies in function approximation, it is reasonable to regard the network as scattering bases to predict scattering coefficients.
Subsequently, it is necessary to embed the physical meanings of the scattering bases into the network and predict the coefficients of different scattering components, making the model simulating the polarimetric decomposition process. To this end, we construct scattering queries and predict the coefficients. Ultimately, it is essential to constrain the foundation model to be pre-trained on large-scale data. To achieve this, a mixed loss function is designed, integrating polarimetric decomposition loss and power self-supervised loss.

\subsection{Network Framework} \label{nframework}

\begin{figure*}[!t]
\centering
\setlength{\abovecaptionskip}{1pt}
\setlength{\belowcaptionskip}{1pt}
\includegraphics[width=0.88\linewidth]{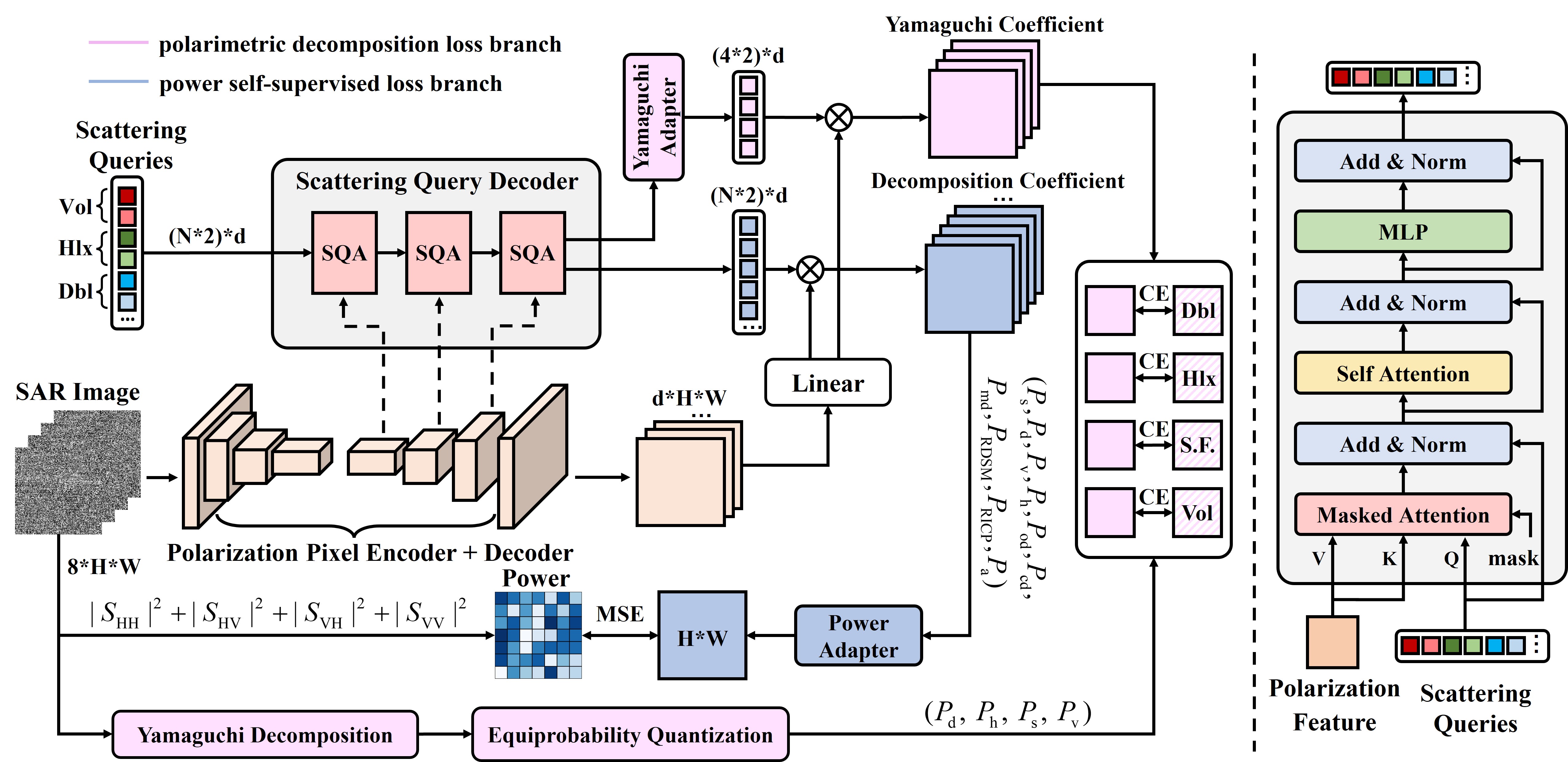}
\caption{\textbf{Overview of our foundation model based on complex-valued SAR images.} Deep network simulates the physical process of polarimetric decomposition, constrained by the polarimetric decomposition loss and power self-supervised loss during the pre-training phase. The embedding and enhancement of the physical meaning of the scattering bases is achieved through scattering query construction.}
\label{framework}
\end{figure*}


The network framework of the foundation model, as illustrated in Fig. \ref{framework}, is composed of four main components.

\noindent \textbf{1) Polarization Pixel Encoder-Decoder (PPED):} The input to the PPED is a combined complex-valued fully PolSAR image, comprising the real and imaginary parts of the HH, HV, VH, and VV modes, resulting in 8 channels. Then, the input image undergoes encoding by the PPED, producing multi-level SAR features that interact with the scattering query decoder. The highest level SAR feature, with dimensions of $d$$\times$$H$$\times$$W$, is subsequently processed to be multiplied with the two-branch refined scatter queries.

\noindent \textbf{2) Scattering Queries:} Scatter queries consist of a series of independent and physically meaningful feature embeddings, constructed in two stages: scatter query initialization and scatter query adaptation (Sec.\ref{sq}). The scatter bases in matrix form are vectorized to initialize the scattering queries. Subsequently, the scatter queries are refined through a dedicated scatter query decoder, adaptively enhancing the semantics to produce enhanced scatter queries across two branches.

\noindent \textbf{3) Scattering Query Decoder (SQD):} The SQD is composed of cascaded scattering query attention (SQA). It takes polarization SAR features and scatter queries as inputs and output scatter queries with dimensions of $(4*2)*d$ and $(N*2*d)$ respectively. The two-branch scatter queries are then multiplied to obtain the Yamaguchi coefficients and decomposition coefficients. As illustrated in Fig. \ref{framework}, the SQA layer is centered around masked attention and self-attention mechanisms. Masked attention selectively focuses on specific regions or aspects of the input, and self-attention captures internal dependencies in scatter queries. Together, these mechanisms enable the SQA layer to generate refined scatter queries that are both semantically meaningful and tailored to the characteristics of the input SAR data.

\noindent \textbf{4) Self-supervised Loss:} The pre-training process is guided by polarimetric decomposition loss and power self-supervision loss. The former compares the Yamaguchi coefficients predicted by the network with ground truth coefficients obtained through Yamaguchi decomposition and equiprobability quantization. The latter ensures power conservation before and after decomposition by comparing the pixel power values calculated from the decomposition coefficients with the input image power. For more details, refer to Sec.\ref{loss}.



\subsection{Scattering Query Construction} \label{sq}
Constructing vector representations based on physical meanings of bases enables better understanding of feature elements. This process includes scattering query initialization and scattering query adaptation.

\subsubsection{Scattering Query Initialization}
Inspired by the seven-component polarimetric scattering decomposition\cite{C7D_2019,C7D_2023}, and the eight-component decomposition technique\cite{C8D_2022}, we construct nine scattering bases with corresponding physical meanings and an adaptive scattering basis. Specifically, the ten scattering bases are defined as: surface scattering $[T_\mathrm{s}]$, double-bounce scattering $[T_\mathrm{d}]$, volume scattering $[T_\mathrm{v}]$, helix scattering $[T_\mathrm{h}]$, oriented dipole scattering $[T_\mathrm{od}]$, compound dipole scattering $[T_\mathrm{cd}]$, mixed dipole scattering $[T_\mathrm{md}]$, rotated dihedral scattering $[T_\mathrm{RDSM}]$, roll-invariant cross polarization scattering $[T_\mathrm{RICP}]$ and adaptive scattering $[T_\mathrm{a}]$. The formula for decile decomposition is:

\begin{align}
[T] &= P_\mathrm{s}[T]_\mathrm{s} + P_\mathrm{d}[T]_\mathrm{d} + P_\mathrm{v}[T]_\mathrm{v} + P_\mathrm{h}[T]_\mathrm{h} + P_\mathrm{od}[T]_\mathrm{od} \notag \\
&\quad + P_\mathrm{cd}[T]_\mathrm{cd} + P_\mathrm{md}[T]_\mathrm{md} + P_\mathrm{RDSM}[T]_\mathrm{RDSM} \notag \\
&\quad + P_\mathrm{RICP}[T]_\mathrm{RICP} + P_\mathrm{a}[T]_\mathrm{a}
\label{decomposition10}
\end{align}

\begin{figure*}[!tp]
\centering
\setlength{\abovecaptionskip}{1pt}
\setlength{\belowcaptionskip}{1pt}
\includegraphics[width=1.0\linewidth]{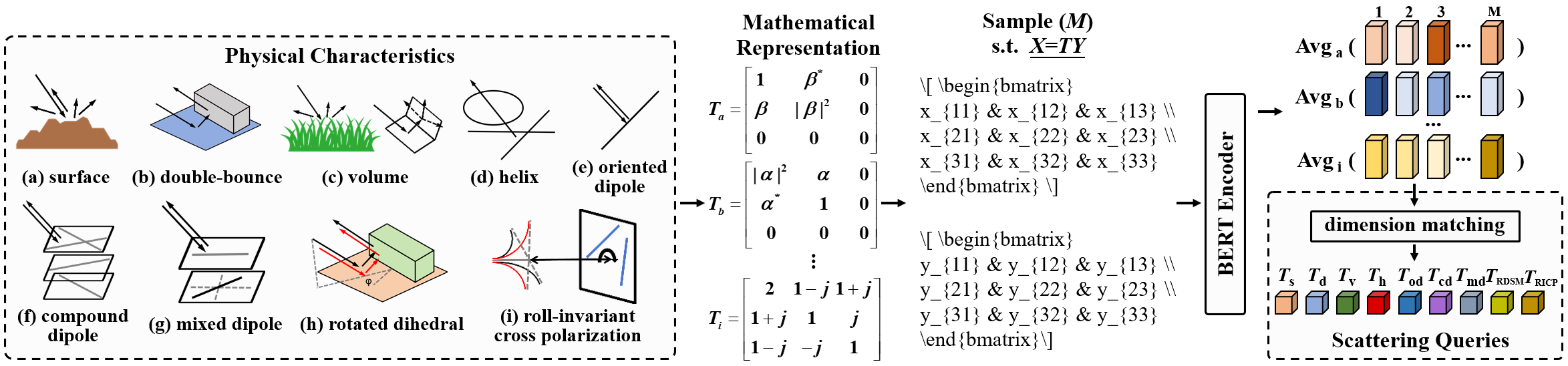}
\caption{\textbf{The process of scattering query initialization.} Nine physical characteristics are represented by corresponding scattering mathematical matrices. Samples are generated based on $X=TY$ and converted into feature vectors by BERT. Scattering queries are then obtained through averaging and matching.
}
\label{Polarization_Basis_Construction}
\end{figure*}

\begin{figure}[!t]
\centering
\setlength{\abovecaptionskip}{1pt}
\setlength{\belowcaptionskip}{1pt}
\includegraphics[width=0.85\linewidth]{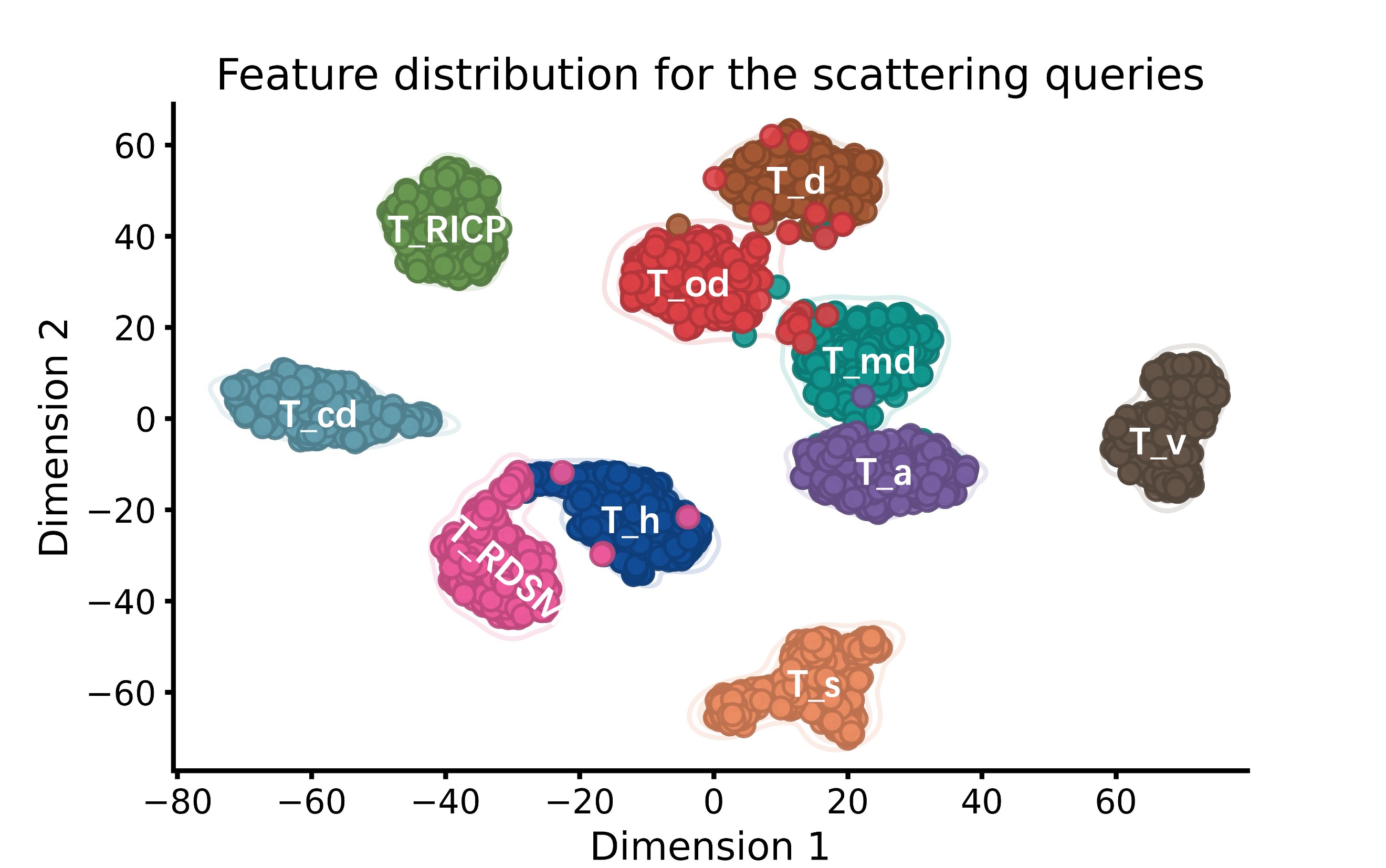}
\caption{\textbf{Feature distribution for the scattering queries.} The features are dimensionality reduced through t-SNE and are independent of each other. }
\label{tsne}
\end{figure}

The process of scattering query initialization is shown in Fig. \ref{Polarization_Basis_Construction}. Excluding adaptive scattering, the generation of scattering queries based on the nine physical characteristics is as follows. These characteristics are represented by corresponding scattering mathematical matrices (i.e. scattering bases). For different bases, $M$ sample pairs satisfying $X=TY$ are randomly generated and encoded into feature vectors using the BERT model\cite{bert}. Then, the feature vectors on $M$ samples are averaged to produce a 1$\times$768-dimensional representation, which represents the semantics of individual bases. Based on the nine bases, a total of 9$\times$256-dimensional scattering queries are finally obtained through matching operation.

As shown in Fig. \ref{tsne}, the scattering queries are validated by the t-SNE dimension reduction. In addition to the nine bases, randomly generated adaptive scattering query is employed to supplement the remaining power. The results indicate that the scattering query vectors can effectively represent the semantics of each scattering component while remaining independent of one another. This establishes meaningful constraints for dividing pixels into specific scattering components.  Ultimately, the network predicts the scattering coefficients corresponding to the bases, ensuring a comprehensive representation.

\subsubsection{Scattering Query Adaptation}
The physical scattering bases are vectorized into scattering queries and further enhanced. As shown in Fig. \ref{framework}, these scattering queries are processed through a scattering query decoder to produce semantically enhanced queries, with dimensions of $(N \times 2) \times d$, where $N$ represents the number of scattering bases, and $d$ is the vector dimension. The queries are subsequently utilized to construct Yamaguchi coefficients and decomposition coefficients, enabling pixel-level decoding.

On the right side of Fig. \ref{framework}, the scattering query decoder operates based on a series of attention layers, with its core utilizing masked attention instead of cross attention. Specifically, each scattering query is associated with a corresponding scattering mechanism, targeting the image regions linked to that mechanism, that is, the segmentation mask $\mathbf{M}$ of a particular scattering component. Cross attention is applied within the masked regions, guiding the network to concentrate on local scattering and accelerating convergence. Subsequently, the self-attention layer captures the global context.

The calculation of masked attention is detailed as follows:
\begin{equation}
\mathbf{W}_l=\operatorname{softmax}\left(\mathbf{M}_{l-1}+\mathbf{Q}_l \mathbf{K}_l^{\mathrm{T}}\right) \mathbf{V}_l+\mathbf{W}_{l-1}
\end{equation}
where $\mathbf{V}$ and $\mathbf{K}$ are derived from the polarization pixel decoder and are multi-scale features extracted from complex-valued SAR images. The scattering query decoder layer facilitates the interaction between image features and scattering features, enhancing the decoding of scattering information across ten scattering components.

\subsection{Self-supervised Loss} \label{loss}
The self-supervised loss includes two parts: polarimetric decomposition loss and power self-supervised loss.
\subsubsection{Polarimetric Decomposition Loss}
To enable the network to better understand the amplitude and phase information in complex-valued SAR images and achieve high interpretability in result extractions, a specialized polarimetric decomposition loss is designed. The function measures the difference between the four components obtained from the original PolSAR image using Yamaguchi decomposition and the corresponding four components predicted by the network. By minimizing this loss, the network can be effectively trained to not only reproduce the numerical values of the components, but also capture the underlying physical features.

The input image $X \in \mathbb{R} {^{8 \times H \times W}}$ is decomposed into four components by the Yamaguchi decomposition algorithm\cite{Yamaguchi2005four}: S.F. values $Y_1$, Dbl values $Y_2$, Vol values $Y_3$, and Hlx values $Y_4$. Since the scattering values typically differ for each pixel in an image, fitting each pixel individually would greatly increase the computational burden and risk overfitting. Therefore, based on the distribution characteristics of the scattering values, we have chosen an appropriate threshold to binarize them, enhancing the learning efficiency of the network.

\begin{figure}[!t]
\centering
\setlength{\abovecaptionskip}{1pt}
\setlength{\belowcaptionskip}{1pt}
\includegraphics[width=1.0\linewidth]{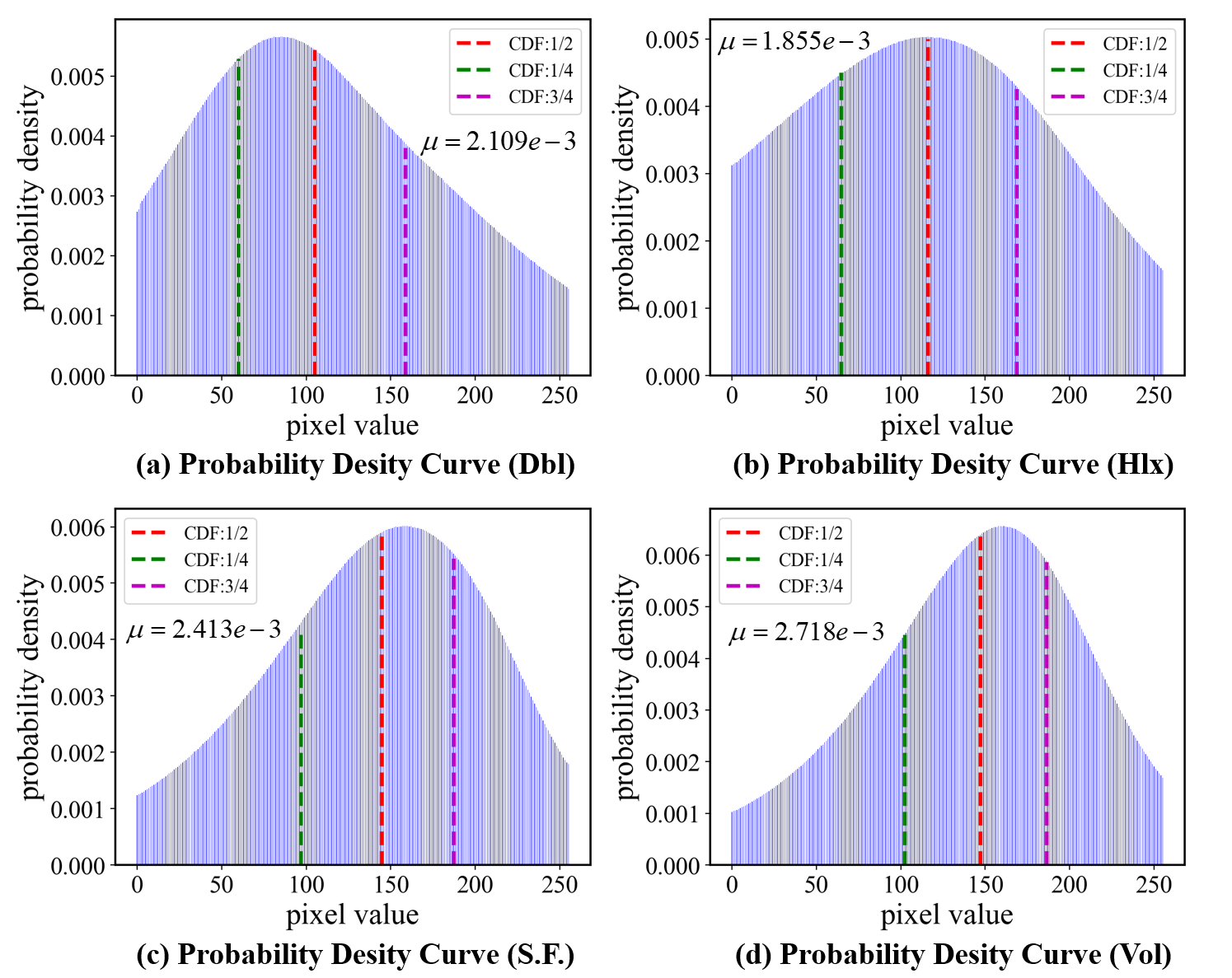}
\caption{\textbf{Statistical distribution of four Yamaguchi scattering components.} Each subgraph provides cumulative density function located at 1/2, 1/4, and 3/4 and scale parameter $\mu$ for fitting the Rayleigh distribution.}
\label{Probability_Density_Curve}
\end{figure}

Firstly, probability density statistics are performed on the scattering values to obtain their probability distributions. Fig. \ref{Probability_Density_Curve} plots the four probability density distributions for the four scattering values, all showing characteristics consistent with the Rayleigh distribution. The probability density function of the Rayleigh distribution is defined as:
\begin{equation}
    f(x,\mu ) = \frac{x}{{{\mu ^2}}}{e^{ - {x^2}/(2{\mu ^2})}},\ x > 0
\end{equation}
where $\mu$ is the scale parameter of the distribution.

Secondly, the scattering value corresponding to a probability value of 0.5 is found and used as the threshold value. Fig. \ref{Probability_Density_Curve} shows that the shape of the probability density curve for the four scattering values is usually skewed, meaning that their distribution is not symmetric. In this case, a probability value of 0.5 divides the data into two equal parts and is a documented reference point in the complex distribution.

Finally, the scattering values are bisected (i.e., divided between 0 and 1) based on the threshold value.

\begin{equation}
    Y_i^j = \left\{ \begin{array}{l}
0,\ \mathrm{if}\ Y_i^j < {\theta _i}\\
1,\ \mathrm{others}
\end{array} \right.
\end{equation}
where $Y_{i}^{j}$denotes the i-th component of the j-th element. $\theta_i$ denotes the value corresponding to a cumulative probability of 0.5 for the i-th component's distribution.


A binary cross-entropy (CE) loss is used to measure the inconsistency between the Yamaguchi decomposition quaternion of the original SAR image and the quaternion predicted by the network to guide the learning process of the network.

\begin{equation}
    \scalebox{0.88}{%
    $ {L_{yamaguchi}} = \sum\limits_{i = 1}^4 {\sum\limits_{j = 1}^{H \times W} {Y_i^j\log R_i^j + \left( {1 - Y_i^j} \right)\log \left( {1 - R_i^j} \right)} } $
    }
\end{equation}
where $R_i^j$ is the network output and represents the i-th component prediction value of the j-th element.

\subsubsection{Power Self-supervised Loss}
With the advancement in high-resolution imaging technology, more accurate scattering modeling is needed. Based on the physical scattering model of incoherent decomposition\cite{C7D_2019,C7D_2023,C8D_2022}, we construct generalized scatterers, and determine the parameters for each component, namely the decomposition coefficients. The ten scattering powers include surface scattering $(P_\mathrm{s})$, double-bounce scattering $(P_\mathrm{d})$, volume scattering $(P_\mathrm{v})$, helix scattering $(P_\mathrm{h})$, oriented dipole scattering $(P_\mathrm{od})$, compound dipole scattering $(P_\mathrm{cd})$, mixed dipole scattering $(P_\mathrm{md})$, rotated dihedral scattering $(P_\mathrm{RDSM})$, roll-invariant cross polarization scattering $(P_\mathrm{RICP})$ and adaptive scattering $(P_\mathrm{a})$. Together, these components constitute the total scattering power.
\begin{align}
\mathrm{SPAN} &= P_\mathrm{s} + P_\mathrm{d} + P_\mathrm{v} + P_\mathrm{h} + P_\mathrm{od} + P_\mathrm{cd} + P_\mathrm{md} \notag \\
&\quad + P_\mathrm{RDSM} + P_\mathrm{RICP} + P_\mathrm{a}
\end{align}
Then, a power map is generated through the power adapter module, capturing the power of a single pixel with a size of $H \times W \times 1$. The power derived from the input complex-valued SAR images, is calculated as the total of the squares of the real and imaginary parts across four polarization modes.
\begin{equation}
\operatorname{SPAN}=\left|S_{\mathrm{HH}}\right|^2+\left|S_{\mathrm{HV}}\right|^2+\left|S_{\mathrm{VH}}\right|^2+\left|S_{\mathrm{VV}}\right|^2
\end{equation}

To ensure the total power remains unchanged before and after decomposition, we impose the condition that the sum of the ten scattering powers equals the total power of the input images. To measure the discrepancy between the actual and predicted power, we constrain the network using Mean Squared Error (MSE) loss for self-supervised training.
\begin{equation}
L_{power}=\frac{1}{N} \sum_{i=1}^N\left(\hat{p}_i-p_i\right)^2
\end{equation}
where $N=H*W$, represents the number of image pixels,  $\hat{p}_i$ represents the predicted power of the i-th pixel, and $p_{i}$ represents the true power.

The overall loss is comprised of the polarimetric decomposition loss and the power self-supervised loss.
\begin{equation}
    L = {L_{yamaguchi}} + \alpha {L_{power}}
\end{equation}
Here, $\alpha$ is a constant used to adjust the proportion of the power self-supervised loss in the total loss, and it is set to 0.1 in the experiment.

\section{Experimental Results}
The foundation model is pre-trained on large-scale unlabeled complex-valued SAR data. To evaluate its effectiveness, we conduct extensive experiments across six tasks on both complex-valued SAR datasets (containing amplitude and phase information) and general SAR datasets (containing only amplitude information). 
Sec.\ref{e_pretrain} provides the pre-training details, while Sec.\ref{e_complexsar} and Sec.\ref{e_generalsar} present the results for the complex-valued SAR datasets and the general SAR datasets, respectively.

\subsection{Pre-training Phase} \label{e_pretrain}

To train the foundation model, large-scale complex-valued SAR data is collected. The images are captured by the GF3 satellite in QPSI imaging mode, with an incidence angle ranging from 20 to 41 degrees. The azimuth resolution is 8m, while the range resolution varies between 6m and 9m. The pre-training dataset includes 400,000 images depicting regions across Asia, North America and South America, with varying levels of economic development. The selected scenes primarily consist of urban and airport areas. 79\% of the cities in the dataset are from China, and 21\% are in other countries. 

\begin{figure}[!t]
\centering
\setlength{\abovecaptionskip}{1pt}
\setlength{\belowcaptionskip}{1pt}
\includegraphics[width=0.9\linewidth]{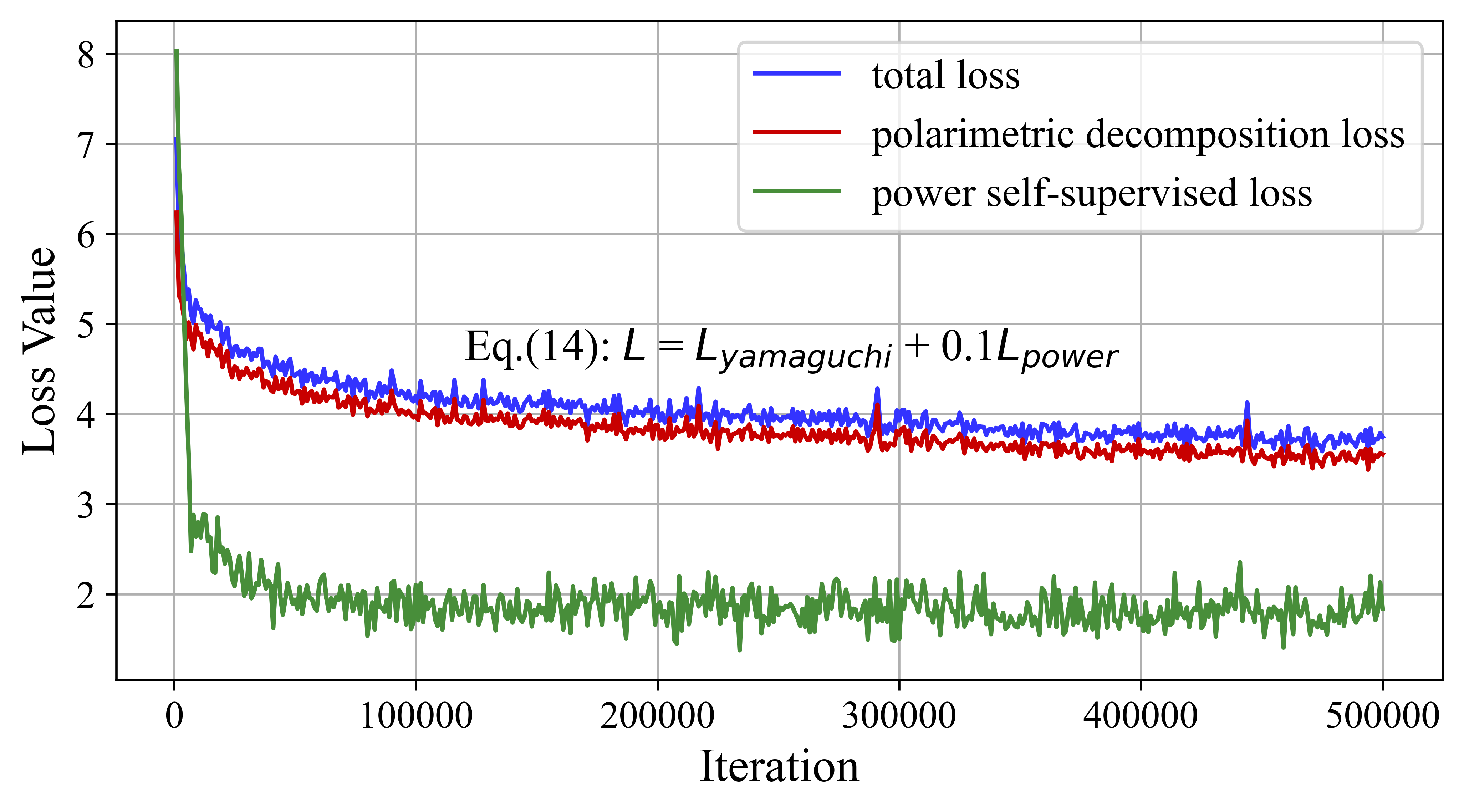}
\caption{\textbf{The loss curve during the pre-training phase.} The total loss, polarimetric decomposition loss and power self-supervised loss steadily decrease to convergence.}
\label{loss_curve}
\end{figure}

\begin{figure}[!t]
\centering
\setlength{\abovecaptionskip}{1pt}
\setlength{\belowcaptionskip}{1pt}
\includegraphics[width=\linewidth]{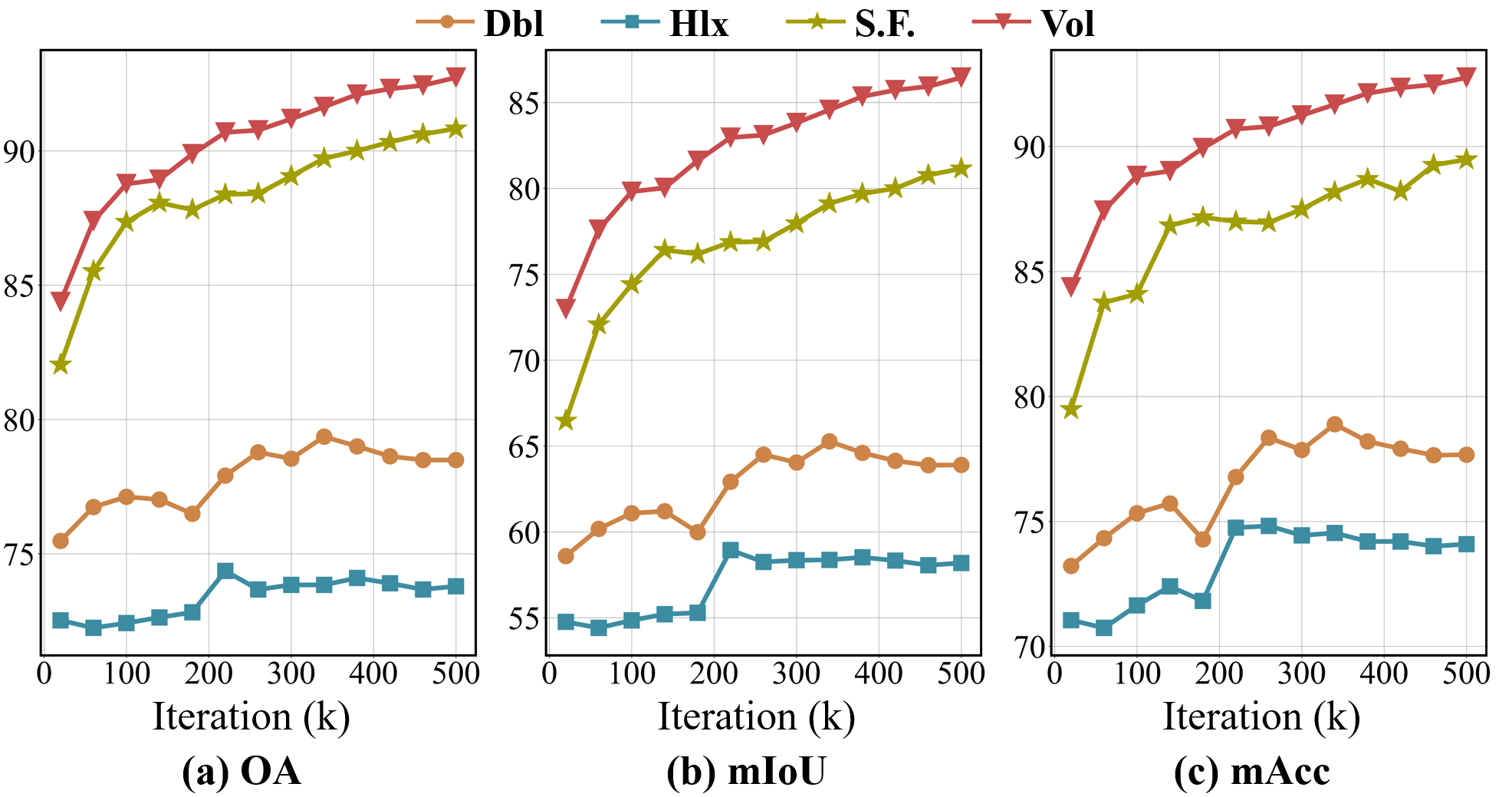}
\caption{\textbf{Quantitative results of polarimetric decomposition coefficients during the pre-training phase.} OA, mIoU, mAcc gradually increase until convergence.}
\label{pretrain_metrics_curve}
\end{figure}

All experiments for the foundation model are conducted using the Pytorch framework with 8 NVIDIA A100 GPUs. Swin-B is used as the backbone, with 8-channel input data comprising the real and imaginary parts. The input image size is 512$\times$512 pixels, with a batch size of 24. An AdamW optimizer is employed, featuring a momentum of 0.9 and a weight decay of 0.05. The basic learning rate is set to 0.008, and a Poly learning strategy with a parameter of 0.9 is adopted. During the pre-training phase, the convergence process of the loss is shown in Fig. \ref{loss_curve}. 

To measure the reconstruction performance of the foundation model, we first analyze multiple polarimetric decomposition coefficients quantitatively, as shown in Fig. \ref{pretrain_metrics_curve}. The prediction performance of the four scattering components gradually stabilizes as iteration increases. Among them, Dbl and Hlx converge rapidly, while S.F. and Vol exhibit a relatively slow convergence. 500k iterations are used to seek the optimal balance. For all four components, the OA exceeds 70\%, indicating that our foundation model maintains a high level of reconstruction performance. Notably, in the Vol and S.F. scattering components, our model demonstrates significant advantages across three metrics, with OA and mAcc surpassing 90\% and mIoU approaching 90\%.

\begin{figure}[!t]
\centering
\setlength{\abovecaptionskip}{1pt}
\setlength{\belowcaptionskip}{1pt}
\includegraphics[width=\linewidth]{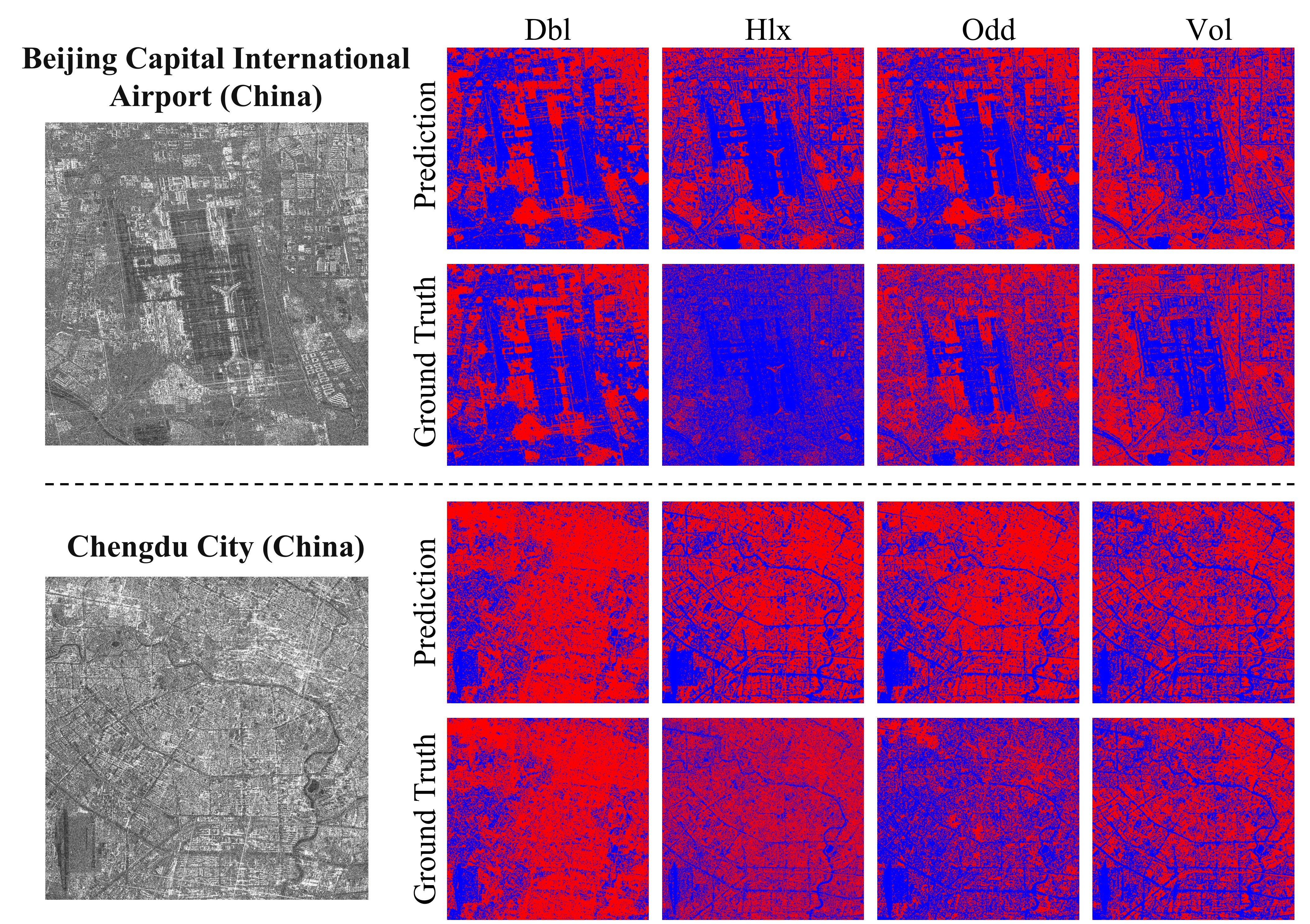}
\caption{\textbf{Scattering reconstruction visualization for the airport and city areas.} The upper line represents the prediction, while the corresponding lower line shows the ground truth. The outline and details can be observed.}
\label{visual_pretrain}
\end{figure}

The reconstructed images of the foundation model are visualized in Fig. \ref{visual_pretrain} to demostrate the reconstruction performance. Beijing Capital International Airport (China) and Chengdu City (China) are selected as representative airport and city. The visualization results reveal that the reconstructed scattering components closely resemble the true values, demonstrating that our method effectively simulates the physical decomposition process of complex-valued SAR images. Additionally, the significant differences between the scattering components align with the theory that each scattering component represents a distinct physical phenomenon. 

To adapt the pre-trained foundation model to downstream tasks, the encoder of the foundation model is used. By leveraging its pre-trained feature representations, the encoder can effectively capture the semantic and scattering characteristics of complex-valued SAR data. This enables efficient fine-tuning for tasks such as semantic segmentation, object detection, and classification, ensuring robust performance even in scenarios with limited labeled data.

\subsection{Performance on Complex-valued SAR Downstream Tasks} \label{e_complexsar}

To explore the superiority of physically inspired representations learned by our foundation model during the pre-training phase, we gradually reduce the reliance on labeled data and conduct experiments on semantic segmentation, few-shot segmentation, and unsupervised classification tasks.

\begin{table*}[!t]
\centering
\setlength{\abovecaptionskip}{1pt}
\setlength{\belowcaptionskip}{1pt}
\caption{Semantic segmentation results on the complex-valued SARSegL1 dataset. Bolded and underlined numbers represent the best and second-best results respectively. The red gains in parentheses indicate the benefits of using our foundation model over the SAR-based foundation model.}
\label{tab_downstream_ss}
\renewcommand{\arraystretch}{1.25}
\resizebox{0.95\linewidth}{!}{
\begin{tabular}{r|c|ccccc|c|c|c}
\toprule[1.0pt] 
\multirow{2}{*}{\textbf{Methods}} & \multirow{2}{*}{\textbf{Pre-trained Model}} & \multicolumn{5}{c|}{\textbf{IoU per category(\%)}} & \multirow{2}{*}{\textbf{mIoU(\%)}} & \multirow{2}{*}{\textbf{mAcc(\%)}} & \multirow{2}{*}{\textbf{OA(\%)}}  \\ 
\cline{3-7} 
\multicolumn{1}{c|}{} & & Water & Vegetation & Bare Land & Road & Building &  &  & \\ 
\midrule[0.7pt]
\midrule[0.7pt]
\multicolumn{1}{r|}{PSPNet(2017)\cite{PSPNet}} & \multirow{9}{*}{Swin-ImageNet-1K} & 68.24 & 67.63 & 51.72 & 13.52 & 65.52 & 53.33 & 64.22 & 76.15 \\ 
\multicolumn{1}{r|}{UperNet(2018)\cite{UperNet}} & {} & 73.00 & \underline{71.57} & \textbf{54.14} & 19.28 & 68.41 & \underline{57.28} & \underline{68.05} & \underline{78.94} \\ 
\multicolumn{1}{r|}{DeepLabV3+(2018)\cite{Deeplabv3+}} & {} & 62.07 & 61.95 & 28.87 & 2.66 & 60.44 & 43.20 & 55.97 & 72.04 \\ 
\multicolumn{1}{r|}{ANN(2019)\cite{ANN}} & {} & 59.30 & 57.90 & 26.80 & 6.56 & 56.59 & 41.44 & 53.66 & 68.50 \\ 
\multicolumn{1}{r|}{CCNet(2019)\cite{CCNet}} & {} & 67.21 & 66.21 & 51.60 & 11.99 & 65.87 & 52.57 & 63.86 & 75.94 \\ 
\multicolumn{1}{r|}{KNet(2021)\cite{KNet}} & {} & 65.92 & 66.56 & 47.14 & 10.63 & 65.08 & 51.06 & 61.52 & 75.30 \\ 
\multicolumn{1}{r|}{SegNeXT(2022)\cite{SegNeXt}} & {} & 64.97 & 63.48 & 39.40 & 4.05 & 63.79 & 47.14 & 57.58 & 73.96 \\ 
\multicolumn{1}{r|}{Mask2former(2022)\cite{mask2former}} & {} & 67.78 & 63.89 & 34.33 & 12.05 & 63.61 & 48.34 & 58.62 & 74.40 \\ 
\multicolumn{1}{r|}{DDRNet(2023)\cite{DDRNet}} & {} & 56.54 & 56.96 & 25.21 & 7.67 & 55.52 & 40.38 & 53.23 & 67.30 \\ 
\midrule[0.7pt]

\multicolumn{1}{r|}{PSPNet(2017)\cite{PSPNet}} & \multirow{3}{*}{\parbox[c]{2cm}{\centering RingMo\cite{RSpretrained2}\\(Optical+SAR)}} & 46.56 & 53.57 & 8.14 & 6.28 & 52.66 & 33.44 & 46.39 & 64.07 \\ 
\multicolumn{1}{r|}{UperNet(2018)\cite{UperNet}} & {} & 73.26 & 68.31 & 43.58 & 5.35 & 65.69 & 51.24 & 63.53 & 76.85 \\ 
\multicolumn{1}{r|}{Mask2former(2022)\cite{mask2former}} & {} & 54.69 & 54.64 & 4.25 & 3.88 & 53.55 & 34.20 & 46.13 & 66.14 \\ 
\midrule[0.7pt]

\multicolumn{1}{r|}{PSPNet(2017)\cite{PSPNet}} & \multirow{3}{*}{\parbox[c]{2cm}{\centering FGMAE\cite{FGMAE}\\(SAR)}} & 61.59 & 51.78 & 13.12 & 3.48 & 55.63 & 37.12 & 49.26 & 66.83 \\ 
\multicolumn{1}{r|}{UperNet(2018)\cite{UperNet}} & {} & 56.73 & 49.91 & 1.90 & 7.34 & 52.86 & 33.74 & 44.51 & 63.30 \\ 
\multicolumn{1}{r|}{Mask2former(2022)\cite{mask2former}} & {} & 62.16 & 56.52 & 1.06 & 5.30 & 56.40 & 36.29 & 47.20 & 68.27 \\ 
\midrule[0.7pt]

\multicolumn{1}{r|}{PSPNet(2017)\cite{PSPNet}} & \multirow{3}{*}{\textbf{Ours}} & 64.90 & 70.36 & 47.28 & 16.17 & \underline{68.64} & 53.47$_{\textcolor{red}{(+16.35)}}$ & 64.61$_{\textcolor{red}{(+15.35)}}$ & 77.64$_{\textcolor{red}{(+10.81)}}$ \\ 
\multicolumn{1}{r|}{UperNet(2018)\cite{UperNet}} & {} & \textbf{84.47} & \textbf{72.87} & \underline{52.52} & \textbf{27.92} & \textbf{69.93} & \textbf{61.52}$_{\textcolor{red}{(+27.78)}}$ & \textbf{70.69}$_{\textcolor{red}{(+26.18)}}$ & \textbf{81.33}$_{\textcolor{red}{(+18.03)}}$ \\ 
\multicolumn{1}{r|}{Mask2former(2022)\cite{mask2former}} & {} & \underline{74.87} & 65.67 & 22.68 & \underline{23.09} & 64.93 & 50.24$_{\textcolor{red}{(+13.95)}}$ & 59.76$_{\textcolor{red}{(+12.56)}}$ & 76.66$_{\textcolor{red}{(+8.39)}}$ \\ 
\bottomrule[1.0pt] 
\end{tabular}}
\end{table*}

\noindent \textbf{1. Datasets.}

1) \textbf{SARSegL1} dataset contains two complex-valued SAR images from the GF3 satellite, captured over Guangzhou and Hangzhou, China, with a 5m resolution. The images have dimensions of 5456$\times$4708 and 6192$\times$4888 pixels, and five pixel-level annotation categories: Water, Vegetation, Bare Land, Road, and Building. It is used for semantic segmentation task and few-shot segmentation task.

2) \textbf{SARClsL1} dataset includes two complex-valued SAR images from the GF3 satellite, with an 8m resolution. The 2018 San Francisco image (5829$\times$7173 pixels) has six classes: Water, Low-density Urban Area (LD.U), High-density Urban Area (HD.U), Developed Area (DA), Mountain, and Vegetation. The 2019 Phra Nakhon Si Ayutthaya image (7525$\times$8356 pixels) has six classes: Building, Road, Grass, Water, Forest, and Land. It is used for unsupervised classification task.

\noindent \textbf{2. Implementation Details.}

For the SARSegL1 and SARClsL1 datasets, the real and imaginary parts of the four modes (HH, HV, VH, and VV) are concatenated to form 8-channel input samples. For semantic segmentation, the input size is set to 500$\times$500 pixels with a batch size of 8. The AdamW optimizer is employed with an initial learning rate of 0.001. The learning rate follows the linear scheduler and the polynomial scheduler. For few-shot segmentation, considering that there are limited examples of class `Road' in SARSegL1, we choose the remaining four classes for few-shot experiments. The SGD optimizer with a momentum of 0.9 and an initial learning rate of 0.005 is employed, with a batch size of 8. For unsupervised classification, the foundation models including Swin-ImageNet-1K, RingMo and FGMAE are used as the backbone to extract physical representations without fine-tuning.


\noindent \textbf{3. Results.}

1) \textbf{Semantic Segmentation}. This task assigns a specific class label to each pixel in an image, aiming to demonstrate the capability of our foundation model to accurately interpret the scene. TABLE \ref{tab_downstream_ss} presents the semantic segmentation results based on different pre-trained foundation models. As shown in the final section of TABLE \ref{tab_downstream_ss}, experimental results on three representative advanced networks reveal that our model possesses significant advantages. Notably, UperNet equipped with our foundation model achieves optimal performance among all methods, with scores of 61.52\%, 70.69\%, and 81.33\% for the mIoU, mAcc, and OA metrics.
Fig.\ref{downstream_seg} shows that our foundation model exhibits significant advantages in terms of road, bare land, water, and vegetation, respectively.

\begin{table*}[!t]
\setlength{\abovecaptionskip}{1pt}
\setlength{\belowcaptionskip}{1pt}
\caption{Performance of Few-shot segmentation task on the SARSegL1 dataset measured in mIoU and FB-IoU (\%) under the 1-shot setting. `Mean' and `FB-IoU' denote the averaged mIoU score and FB-IoU score for all classes, respectively.} \label{Performance of FSS}
\renewcommand\arraystretch{1.25}
\centering
\resizebox{0.7\linewidth}{!}{
\begin{tabular}{c|cc|c|c|cc} 
\toprule[1.0pt]
\multirow{2}{*}{\textbf{Pre-trained Model}} & \multicolumn{2}{c}{\textbf{Fold-0}}      & \multicolumn{1}{c}{\textbf{Fold-1}} & \textbf{Fold-2}        & \multirow{2}{*}{\textbf{Mean IoU(\%)}} & \multirow{2}{*}{\textbf{FB-IoU(\%)}}  \\ 
\cline{2-5}
                                   & Water          & Vegetation     & Bare land                  & Building       &                       &                          \\ 
\midrule[0.7pt]
\midrule[0.7pt]
ResNet-50-ImageNet-1K              & 14.71          & 28.45          & 6.53                       & 28.66          & 19.59                 & 35.94                    \\
Swin-ImageNet-1K                   & 13.14          & 29.17          & 7.31                       & 30.23          & 19.96                 & 37.36                    \\ 
\midrule[0.7pt]
\textbf{Ours}                      & \textbf{19.32} & \textbf{32.91} & \textbf{7.42}              & \textbf{39.92} & \textbf{24.89}$_{\textcolor{red}{(+4.93)}}$        & \textbf{40.97}$_{\textcolor{red}{(+3.61)}}$           \\
\bottomrule[1.0pt]
\end{tabular}}
\end{table*}

\begin{table}[!t]
\centering
\setlength{\abovecaptionskip}{1pt}
\setlength{\belowcaptionskip}{1pt}
\caption{Split details of dataset classes on the Few-shot segmentation.}
\label{split details in FSS}
\renewcommand{\arraystretch}{1.25}
\resizebox{0.85\linewidth}{!}{
\begin{tabular}{c|c|c} 
\toprule[0.9pt] 
\# Fold & Training classes                       & Testing classes      \\ 
\midrule[0.6pt]
\midrule[0.6pt]
0     & Bare Land, Building             & Water, Vegetation    \\
1     & Water, Vegetation, Building                & Bare Land   \\
2     & Water, Vegetation, Bare Land  & Building                 \\
\bottomrule[0.9pt] 
\end{tabular}}
\end{table}

2) \textbf{Few-shot Segmentation}. This task leverages limited labeled samples to achieve accurate segmentation, with the goal of evaluating the model's generalization ability in data-scarce scenarios. Following the setup of AgMTR \cite{bi2024agmtr, bi2023not}, we divide the four classes in SARSegL1 into three folds, as shown in TABLE \ref{split details in FSS}, with distinct training and testing classes to simulate seen and unseen categories. TABLE \ref{Performance of FSS} compares performance on the few-shot segmentation task under the 1-shot setting using various pre-trained foundation models. Results show that AgMTR, utilizing our model for feature extraction, achieves the best performance, outperforming other models by 4.93\% in mIoU and 3.61\% in FB-IoU. This highlights that the proposed foundation model is able to effectively extract features for accurate segmentation, even for unseen classes, demonstrating strong generalization.

3) \textbf{Unsupervised Classification}. This task extracts latent features using the foundation model and performs classification without annotations, aiming to validate the model's generalization and adaptability. With the assurance that the clustering algorithm and post-processing operations remain consistent, TABLE \ref{tab_downstream_Cla} presents the unsupervised classification results on the SARClsL1 dataset. The mAP shows that our physical feature representations achieve excellent results, outperforming the second-best model by 16.6\% and 7.1\% respectively. Fig. \ref{downstream_cla} shows the unsupervised classification visualization results of Yamaguchi algorithm, RingMo, FGMAE, and our foundation model.
This highlights that our foundation model extracts meaningful representations and can achieve satisfactory classification results without fine-tuning.

\begin{table*}[t]
\centering
\setlength{\abovecaptionskip}{1pt}
\setlength{\belowcaptionskip}{1pt}
\caption{Unsupervised classification results on SARClsL1 dataset. Bolded and underlined numbers represent the best and second-best results respectively. The red gains in parentheses indicate the benefits of using our pre-trained foundation model over the second-best foundation model.}
\label{tab_downstream_Cla}
\renewcommand{\arraystretch}{1.25}
\resizebox{0.8\linewidth}{!}{ 
\begin{tabular}{c|c|>{\centering\arraybackslash}p{0.9cm}>{\centering\arraybackslash}p{0.9cm}>{\centering\arraybackslash}p{0.9cm}>{\centering\arraybackslash}p{0.9cm}>{\centering\arraybackslash}p{0.9cm}>{\centering\arraybackslash}p{1.2cm}|c}
\toprule
\multicolumn{1}{c|}{\textbf{Scenario}} & \textbf{Pre-trained Model}  & $\textbf{Water}$  & $\textbf{DA}$  & $\textbf{HD.U}$  & $\textbf{LD.U}$  & $\textbf{Mountain}$  & $\textbf{Vegetation}$ & $\textbf{mAP(\%)}$  \\ 

\midrule
\midrule
\multirow{4}{*}{San Francisco} & Swin-ImageNet-1K  & 94.9 & 2.5 & 68.6 & \textbf{45.7} & \underline{46.3} & 43.2 & \underline{50.2} \\
& RingMo\cite{RSpretrained2}  & \underline{96.4} & 2.1 & \textbf{85.1} & 37.6 & 33.0 & 34.1 & 48.1 \\
& FGMAE\cite{FGMAE}  & 96.0 & \underline{3.2} & 58.6 & 41.0 & 35.5 & \underline{43.6} & 46.3 \\
\cline{2-9}
& \textbf{Ours}  & \textbf{96.5} & \textbf{84.1} & \underline{74.3} & \underline{41.8} & \textbf{55.2} & \textbf{48.7} & \textbf{66.8}$_{\textcolor{red}{(+16.6)}}$ \\

\midrule
\multicolumn{1}{c|}{\textbf{Scenario}} & \textbf{Pre-trained Model}  & $\textbf{Building}$  & $\textbf{Road}$  & $\textbf{Grass}$  & $\textbf{Water}$  & $\textbf{Forest}$  & $\textbf{Land}$  & $\textbf{mAP(\%)}$  \\ 

\midrule
\midrule
\multirow{4}{*}{\begin{tabular}[c]{@{}c@{}}Phra Nakhon\\Si Ayutthaya\end{tabular}} & Swin-ImageNet-1K  & 79.3 & 2.0 & 48.9 & \underline{31.1} & \underline{8.0} & \underline{21.1} & 31.7 \\
& RingMo\cite{RSpretrained2}  & 79.4 & 4.6 & 53.2 & 28.4 & 6.8 & 20.2 & 32.1 \\
& FGMAE\cite{FGMAE}  & \underline{80.9} & \underline{5.0} & \underline{56.6} & 27.4 & 5.9 & 19.9 & \underline{32.6} \\
\cline{2-9}
& \textbf{Ours}  & \textbf{93.2} & \textbf{5.6} & \textbf{64.7} & \textbf{32.0} & \textbf{14.9} & \textbf{22.3} & \textbf{38.8}$_{\textcolor{red}{(+7.1)}}$ \\

\bottomrule 
\end{tabular}}
\end{table*}

\begin{figure}[!t]
\centering
\setlength{\abovecaptionskip}{1pt}
\setlength{\belowcaptionskip}{1pt}
\includegraphics[width=1.0\linewidth]{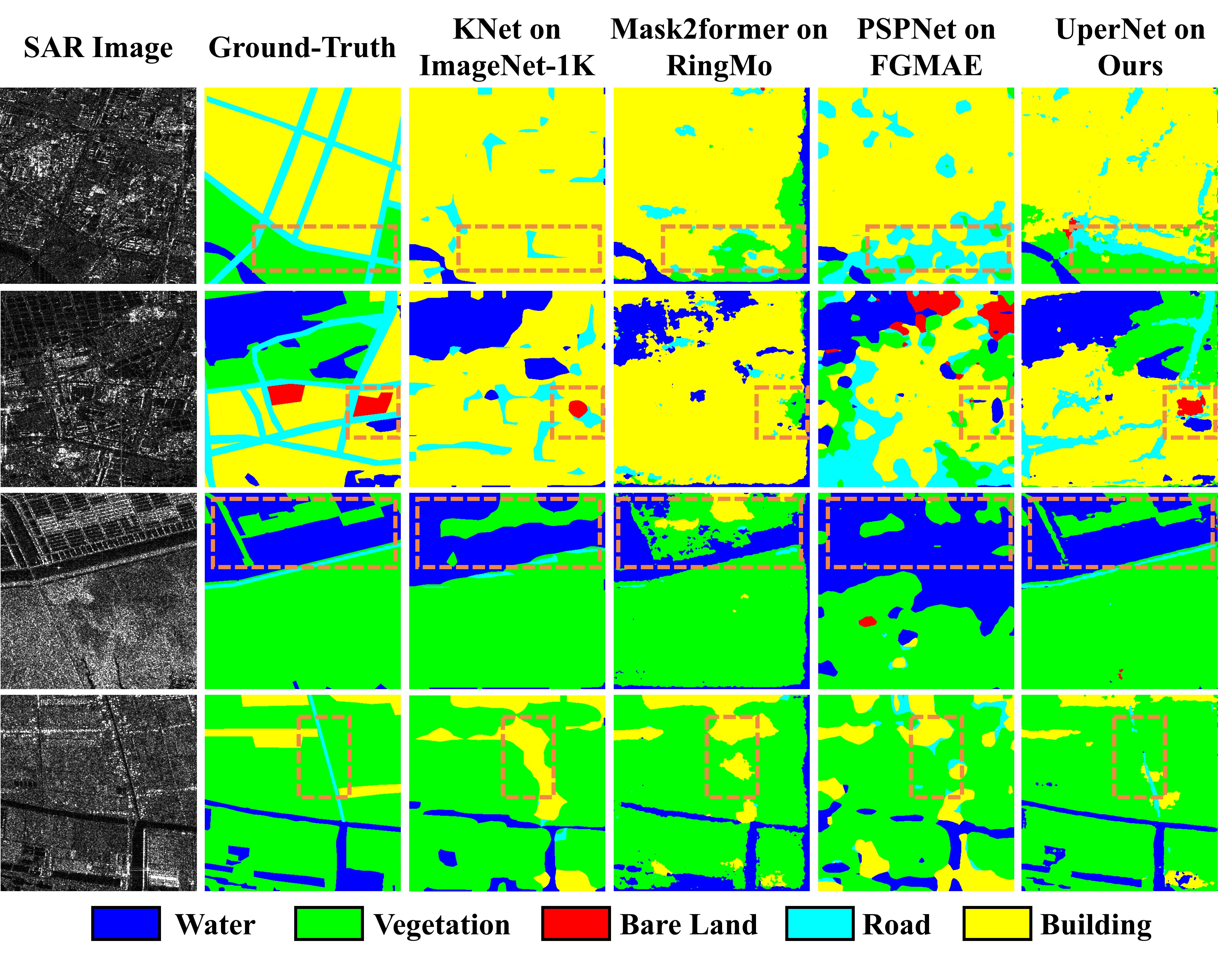}
\caption{\textbf{Visualization results of different foundation models on SARSegL1 dataset.} From the top line to the last line, our model shows significant advantages in terms of road, bare land, water, and vegetation, respectively.}
\label{downstream_seg}
\end{figure}

\begin{figure}[!t]
\centering
\setlength{\abovecaptionskip}{1pt}
\setlength{\belowcaptionskip}{1pt}
\includegraphics[width=1.0\linewidth]{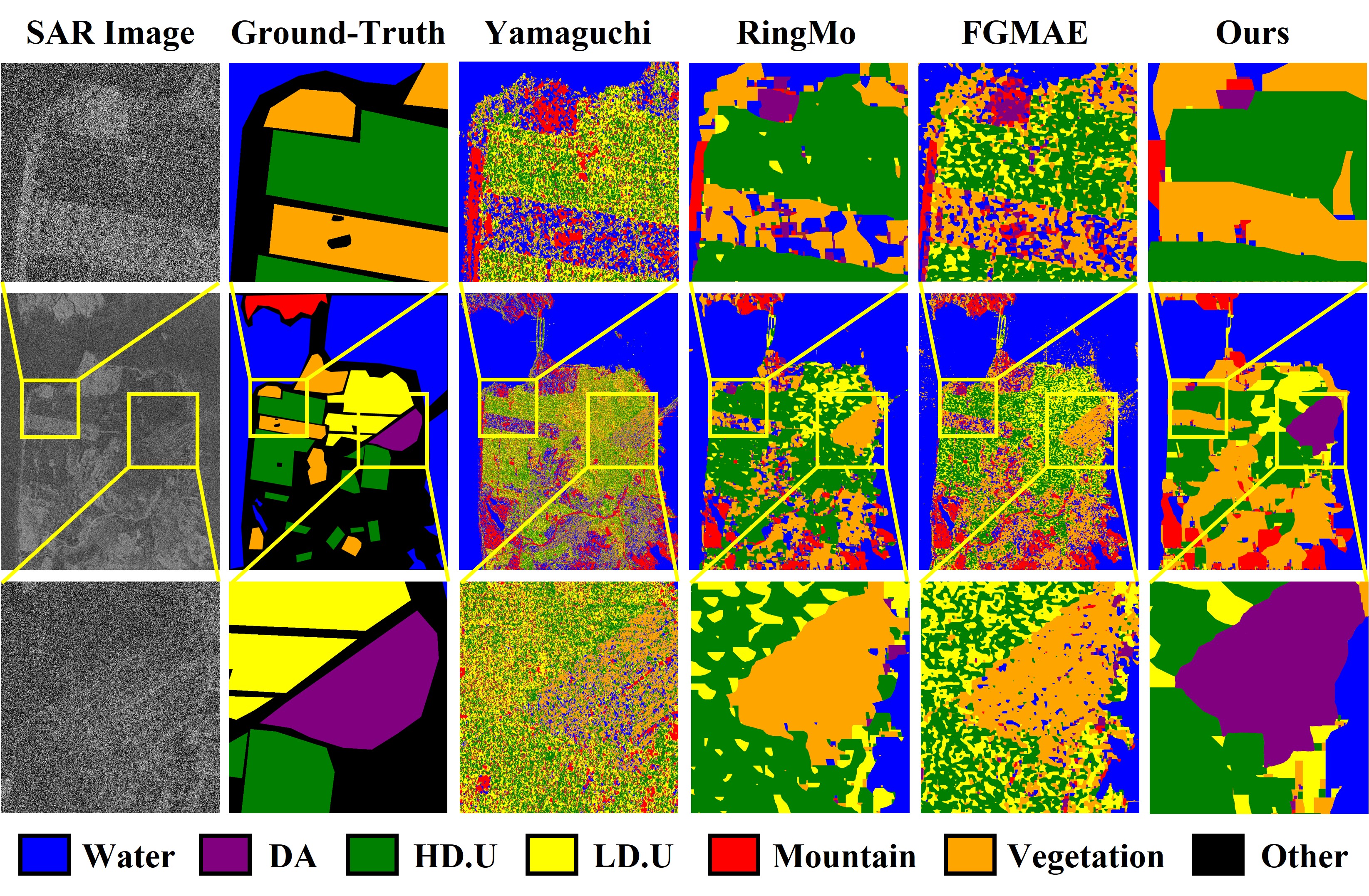}
\caption{\textbf{Unsupervised classification visualization results for the SARClsL1 dataset.} Taking San Francisco scenario as an example, our foundation model has significant advantages without fine-tuning.}
\label{downstream_cla}
\end{figure}

\subsection{Performance on General SAR Downstream Tasks}    \label{e_generalsar}
The foundation model extracts information from both amplitude and phase images and can be generalized to publicly available general SAR datasets (i.e., amplitude images). Downstream tasks including  object detection and semantic segmentation are used for generalization validation.

\noindent \textbf{1. Datasets.}

1) \textbf{HRSID\cite{HRSID}} is a comprehensive dataset for instance segmentation and ship detection in SAR images, formatted following the COCO dataset. It includes 16,951 annotated ship instances across various resolutions (0.5m, 1m, 3m), polarizations, and sea states.

2) \textbf{SAR-AIRcraft-1.0\cite{SAR-AIRcraft-1.0}} is a high-resolution SAR image dataset for aircraft detection, featuring 4,368 images in four sizes (800$\times$800, 1000$\times$1000, 1200$\times$1200, 1500$\times$1500) and 16,463 annotated aircraft targets across seven aircraft categories.

3) \textbf{AIR-PolSAR-Seg\cite{AIR-PolSAR-Seg}} is a semantic segmentation dataset for SAR images, providing pixel-level annotations for six categories: Housing, Industrial, Natural, Land-use Areas, Water, and other regions. It includes 500 SAR images with an 8m resolution, each divided into $512\times512$ pixel blocks.

\noindent \textbf{2. Implementation Details.}

Here are the implementation details of general SAR datasets. For the object detection datasets, including HRSID and SAR-AIRcraft-1.0, all experiments are performed in the mmdetection framework. 
The optimizer utilizes SGD with an initial learning rate of 0.005.
For the semantic segmentation dataset AIR-PolSAR-Seg, the AdamW optimizer is utilized with an initial learning rate of 6e-5 and a weight decay of 0.05. 

\begin{table*}[t]
\centering
\setlength{\abovecaptionskip}{1pt}
\setlength{\belowcaptionskip}{1pt}
\caption{Detection results on the HRSID dataset. Bolded numbers and underlined numbers represent the best and second-best results, respectively. The red gains in parentheses indicate the additional benefits of loading our pre-trained foundation model compared to loading the Swin-ImageNet-1K model.}
\label{tab_downstream_od}
\renewcommand{\arraystretch}{1.25}
\resizebox{0.85\linewidth}{!}{ 
\begin{tabular}{r|c|ccc|ccc}
\toprule
\multicolumn{1}{r|}{\textbf{Methods}} & \textbf{Pre-trained Model}  & \textbf{AP$_{50}$(\%)}  & \textbf{AP$_{50}$(\%)}  & \textbf{AP$_{75}$(\%)}  & \textbf{AP$_{s}$(\%)}  & \textbf{AP$_{m}$(\%)}  & \textbf{AP$_{l}$(\%)}  \\ 
\midrule
\midrule
Mask R-CNN(2017) \cite{maskrcnn} &\multirow{6}{*}{Swin-ImageNet-1K} & 59.5 & 78.2 & 68.4 & 60.3 & 64.6 & 46.5 \\
Cascade R-CNN(2018) \cite{cascadercnn} & & 64.6 & 79.5 & 73.7 & 64.4 & 66.9 & 49.0 \\
TOOD(2021) \cite{TOOD} & & 62.0 & 87.2 & 70.2 & 63.0 & 64.2 & 51.0 \\
Swin-PAFF(2023) \cite{Swin-PAFF} & & 64.6 & 91.3 & 73.3 & 65.7 & 67.9 & 45.7 \\
FBUA-Net(2023) \cite{FBUA} & & \underline{69.1} & 90.3 & \underline{79.6} & \underline{69.6} & 66.4 & 50.2 \\
DDQ DETR(2023) \cite{ddqdetr} & & 67.8 & \underline{92.2} & 78.5 & 69.2 & 67.8 & 52.8 \\ 

\midrule
Mask R-CNN(2017) \cite{maskrcnn} & \multirow{3}{*}{RingMo~\cite{RSpretrained2}} & 52.8 & 79.0 & 61.7 & 55.2 & 51.3 & 10.2 \\
Cascade R-CNN(2018) \cite{cascadercnn} & & 63.3 & 83.6 & 72.4 & 64.6 & 64.0 & 16.5 \\
DDQ DETR(2023) \cite{ddqdetr} & & 63.4 & 89.4 & 73.3 & 65.2 & 61.1 & 29.1 \\ 

\midrule
Mask R-CNN(2017) \cite{maskrcnn} & \multirow{3}{*}{FGMAE\cite{FGMAE}} & 54.5 & 78.8 & 65.7 & 56.9 & 52.4 & 23.4 \\
Cascade R-CNN(2018) \cite{cascadercnn} & & 65.9 & 83.9 & 74.4 & 65.2 & 67.3 & 35.7 \\
DDQ DETR(2023) \cite{ddqdetr} & & 64.8 & 91.5 & 77.6 & 68.3 & 65.7 & 40.1 \\ 

\midrule
Mask R-CNN(2017) \cite{maskrcnn} & \multirow{3}{*}{\textbf{Ours}} & 60.5$_{\textcolor{red}{(+1.0)}}$ & 81.0$_{\textcolor{red}{(+2.8)}}$ & 70.1$_{\textcolor{red}{(+1.7)}}$ & 60.7$_{\textcolor{red}{(+0.4)}}$ & 67.1$_{\textcolor{red}{(+2.5)}}$ & 50.0$_{\textcolor{red}{(+3.5)}}$ \\
Cascade R-CNN(2018) \cite{cascadercnn} & & 66.3$_{\textcolor{red}{(+1.7)}}$ & 84.4$_{\textcolor{red}{(+4.9)}}$ & 76.0$_{\textcolor{red}{(+2.3)}}$ & 67.1$_{\textcolor{red}{(+2.7)}}$ & \textbf{71.2}$_{\textcolor{red}{(+4.3)}}$ & \underline{52.0}$_{\textcolor{red}{(+3.0)}}$ \\
DDQ DETR(2023) \cite{ddqdetr} & & \textbf{70.3}$_{\textcolor{red}{(+2.5)}}$ & \textbf{93.9}$_{\textcolor{red}{(+1.7)}}$ & \textbf{82.3}$_{\textcolor{red}{(+3.8)}}$ & \textbf{72.5}$_{\textcolor{red}{(+3.3)}}$ & \underline{70.0}$_{\textcolor{red}{(+2.2)}}$ & \textbf{56.2}$_{\textcolor{red}{(+3.4)}}$ \\
\bottomrule 
\end{tabular}}
\end{table*}

\begin{table*}[t]
\renewcommand\arraystretch{1.25}
\caption{Performance comparison on SAR aircraft detection task, i.e., SAR-AIRCraft-1.0 datasets.}
\label{sarplane}
\centering
\resizebox{0.85\linewidth}{!}{
\begin{tabular}{r|ccccccc|c} 
\toprule
\multirow{2}{*}{\textbf{Methods}}  & \multicolumn{7}{c|}{\textbf{AP$_{50}$ per category(\%)}}                                                               & \multirow{2}{*}{\textbf{mAP(\%)}}  \\ 
\cline{2-8}
                         & A330          & A320/321      & A220          & ARJ21         & Boeing737     & Boeing787     & Other         &                           \\ 
\midrule
\midrule
Cascade R-CNN(2018)~\cite{cascadercnn}     & 87.4          & 97.5          & 74.0          & 78.0          & 54.5          & 68.3          & 69.1          & 75.7                      \\
YOLOX-Nano(2021)~\cite{ge2021yolox}        & 74.7          & 96.9          & 79.7          & 78.7          & 66.6          & 78.2~         & 73.8          & 81.3                      \\
SA-Net(2021)~\cite{zhirui2023sar}            & 88.6          & 94.3          & 90.3          & 78.6          & 59.7          & 70.8~         & 71.3          & 77.7                      \\
SkG-Net(2022)~\cite{fu2021scattering}           & 66.4          & 78.2          & 66.4          & 65.0          & 65.1          & 69.6~         & 71.4          & 70.7                      \\
YOLOv8s(2023)~\cite{yolov5}                 & 95.2          & 97.7          & \textbf{95.8}  & 86.6          & \underline{78.9}  & 90.9~         & 84.4          & 89.6                      \\
MLSDNet(2023)~\cite{chang2023mlsdnet}           & 91.5          & 96.9          & 85.1          & 83.2          & 71.7          & 72.1          & 78.4          & 82.7                      \\
DiffusionDet(2023)~\cite{chen2023diffusiondet}      & 95.4          & 98.1          & 80.8          & 84.2          & 70.9          & 91.4          & \underline{86.4}  & 86.6                      \\
DiffDet4SAR(2024)~\cite{zhou2024diffdet4sar}       & \textbf{97.1}  & \underline{99.4}  & 82.3          & \underline{87.2}  & 72.8          & 93.3          & 85.9          & 88.4                      \\
SARATR-X(2024)~\cite{yang2024saratr}          & -             & -             & -             & -             & -             & -             & -             & 86.1                      \\
SFS-CNet(2024)~\cite{li2024unleashing}          & 95.9          & 99.3          & 87.9          & 86.7          & 77.9          & \underline{92.9}         & 85.6          & \underline{89.7}              \\ 
\midrule
\textbf{Cascade R-CNN+Ours} & \underline{96.1} & \textbf{99.5} & \underline{94.2} & \textbf{92.3} & \textbf{86.4} & \textbf{95.7} & \textbf{92.1} & \textbf{93.8}             \\
\bottomrule
\end{tabular}}
\end{table*}

\begin{figure}[t]
\centering
\setlength{\abovecaptionskip}{1pt}
\setlength{\belowcaptionskip}{1pt}
\includegraphics[width=1.0\linewidth]{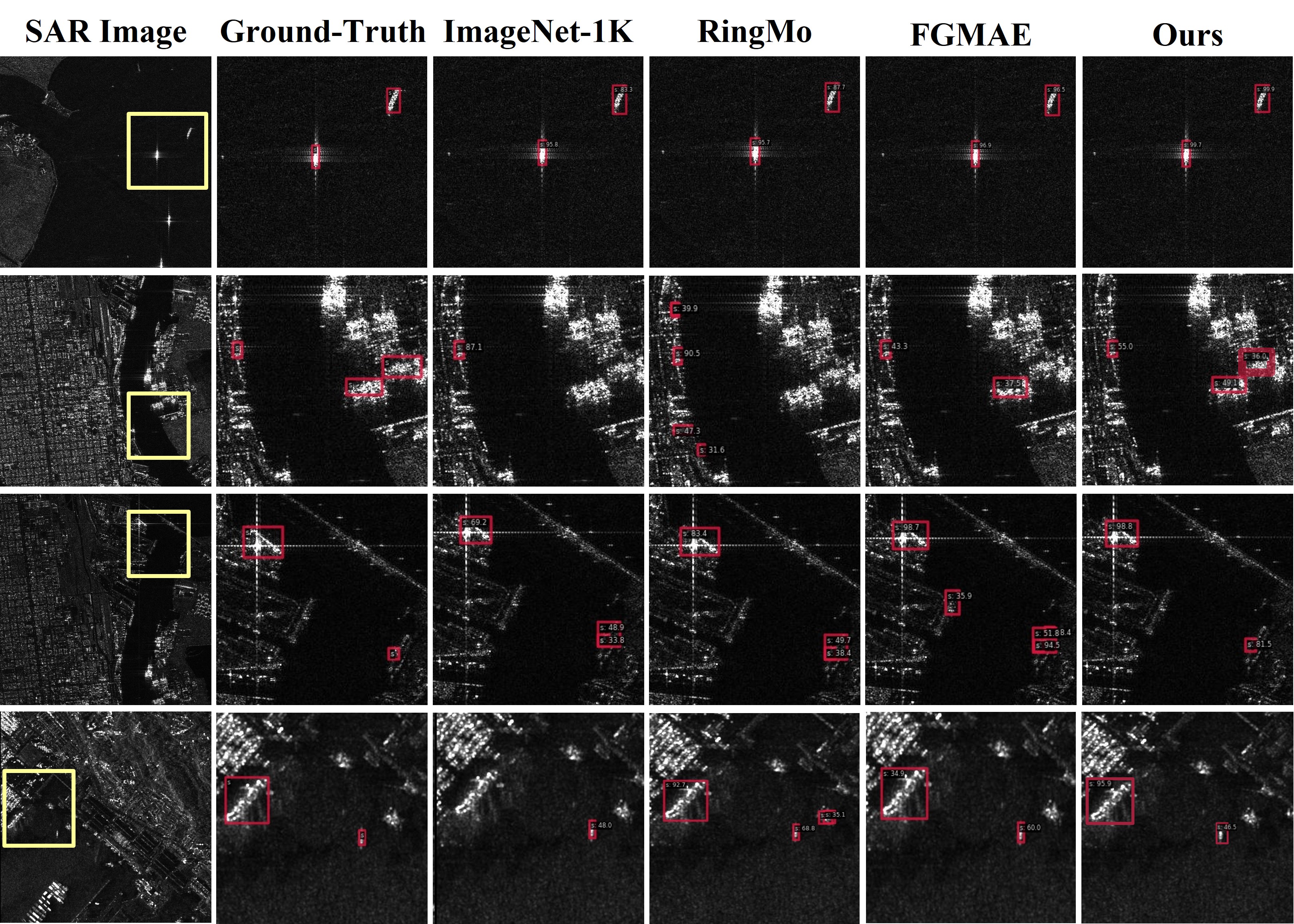}
\caption{\textbf{Visualization of ship detection on the HRSID dataset.} The results of different pre-trained foundation models with DDQ DETR are shown. The last column demonstrates that the detection results align more closely with the ground truth when using our foundation model.}
\label{downstream_dect}
\end{figure}

\begin{table}[t]
\setlength{\abovecaptionskip}{1pt}
\setlength{\belowcaptionskip}{1pt}
\centering
\fontsize{10.5}{12}\selectfont
\renewcommand\arraystretch{1.25}
\caption{Performance comparison on the general SAR semantic segmentation dataset, i.e., AIR-POLSAR-SEG dataset.}
\label{AIR-PoLSAR-SEG}
\resizebox{0.95\linewidth}{!}{
\begin{tabular}{r|cccc|c}
\toprule
\textbf{Methods}   & \renewcommand\arraystretch{1.1}\begin{tabular}[c]{@{}c@{}}\textbf{Industrial} \\\textbf{Area}\end{tabular} & \renewcommand\arraystretch{1.1}\begin{tabular}[c]{@{}c@{}}\textbf{Natural} \\\textbf{Area}\end{tabular} & \textbf{Water} & \textbf{Housing} & \textbf{mIoU(\%)}           \\ 

\midrule
\midrule
NonLocal     \cite{NonLocal}         & 35.51                    & 72.12                 & 70.60                   & 68.39             & 61.66          \\    
EMANet       \cite{EMANet}           & 36.92                    & 72.49                 & 69.72                   & 67.98             & 61.78          \\    
CCNet        \cite{CCNet}            & 32.54                    & 72.27                 & \textbf{72.58}          & 66.75             & \underline{64.04}          \\    

FUSAR-Map    \cite{FUSARmap}         & 38.52                    & 74.09                 & 68.17                   & 62.88             & 60.92          \\    
SegNeXT      \cite{SegNeXt}          & 41.07                    & 69.30                 & 67.37                   & 62.53             & 60.07          \\    
SAFE         \cite{SAFE}             & \underline{44.07}                    & \underline{74.17}                 & 63.09                   & \underline{70.16}             & 62.87          \\    

\midrule
\textbf{UperNet+Ours}                        & \textbf{46.11}           & \textbf{74.22}        & \underline{72.36}                   & \textbf{70.20}    & \textbf{65.72}          \\ 

\bottomrule
\end{tabular}}
\end{table}

\noindent \textbf{3. Results.}

1) \textbf{Ship Detection}. TABLE \ref{tab_downstream_od} presents the results of ship detection on the HRSID dataset, comparing various founation models, including Swin-ImageNet-1K, RingMo, FGMAE and ours. By leveraging features extracted from our foundation model, DDQ DETR consistently outperforms the Swin-ImageNet-1K across all metrics, with improvements of 2.5\%, 1.7\%, 3.8\%, 3.3\%, 2.2\%, and 3.4\% in AP, AP$_{50}$, AP$_{75}$, AP$_{s}$, AP$_{m}$, and AP$_{l}$, respectively. Fig. \ref{downstream_dect} provides visualization results of four cropped regions, illustrating that our model yields fewer false positives and false negatives and aligns more closely with the ground truth. The experimental results demonstrate that our foundation model, pre-trained on complex-valued SAR data, effectively extracts comprehensive and meaningful target scattering information, achieving strong generalization performance on ship detection task that rely solely on amplitude information.

2) \textbf{Aircraft Detection}. TABLE \ref{sarplane} illustrates the experimental results for the fine-grained aircraft detection task. Cascade R-CNN\cite{cascadercnn} is used as the detection head. Our model demonstrates notable advantages, achieving the best performance in 5 out of 7 categories and surpassing the second-best method by 4.1\% in mAP. These results highlight that the significance of physically inspired foundation model, offering innovative strategies to enhance model performance and generalization.

3) \textbf{Semantic Segmentation}. TABLE \ref{AIR-PoLSAR-SEG} presents the results for the general SAR segmentation dataset AIR-PolSAR-Seg. UperHead\cite{UperNet} is employed as the decoder head. Notably, our foundation model achieves state-of-the-art performance, surpassing the second best mIoU by 1.68\%. In terms of land cover categories, it reaches 46.11\% in industrial area, 74.22\% in natural area, 72.36\% in water, and 70.20\% in housing. The results demonstrate that the pre-trained foundation model for complex-valued SAR data generalizes effectively to the semantic segmentation task of general SAR data.

\section{Conclusion}
To address the current limitations in SAR deep learning methods, such as information underutilization and poor interpretability, this paper proposes a remote sensing foundation model for complex-valued SAR images. By simulating the physical process of SAR polarimetric decomposition, the model provides the deep network with physical interpretability. Extensive experiments across six downstream tasks, using both complex-valued and general SAR datasets, validate the effectiveness of the proposed foundation model, significantly improving the interpretation performance on SAR tasks. In summary, our work provides a new perspective for SAR interpretation. In the future, we aim to develop a more comprehensive, multi-modal foundation model to better support the remote sensing community.

\bibliographystyle{IEEEtran}

\end{document}